# Cutset Sampling for Bayesian Networks

**Bozhena Bidyuk**                                                    BBIDYUK@ICS.UCI.EDU
**Rina Dechter**                                                      DECHTER@ICS.UCI.EDU
*School of Information and Computer Science*
*University Of California Irvine*
*Irvine, CA 92697-3425*

## Abstract

The paper presents a new sampling methodology for Bayesian networks that samples only a subset of variables and applies exact inference to the rest. Cutset sampling is a network structure-exploiting application of the Rao-Blackwellisation principle to sampling in Bayesian networks. It improves convergence by exploiting memory-based inference algorithms. It can also be viewed as an anytime approximation of the exact cutset-conditioning algorithm developed by Pearl. Cutset sampling can be implemented efficiently when the sampled variables constitute a loop-cutset of the Bayesian network and, more generally, when the induced width of the network's graph conditioned on the observed sampled variables is bounded by a constant $w$. We demonstrate empirically the benefit of this scheme on a range of benchmarks.

## 1. Introduction

Sampling is a common method for approximate inference in Bayesian networks. When exact algorithms are impractical due to prohibitive time and memory demands, it is often the only feasible approach that offers performance guarantees. Given a Bayesian network over the variables $X = \{X_1, ..., X_n\}$, evidence $e$, and a set of samples $\{x^{(t)}\}$ from $P(X|e)$, an estimate $\hat{f}(X)$ of the expected value of function $f(X)$ can be obtained from the generated samples via the *ergodic average*:

$$E[f(X)|e] \approx \hat{f}(X) = \frac{1}{T} \sum_t f(x^{(t)}) \ , \tag{1}$$

where $T$ is the number of samples. $\hat{f}(X)$ can be shown to converge to the exact value as $T$ increases. The central query of interest over Bayesian networks is computing the posterior marginals $P(x_i|e)$ for each value $x_i$ of variable $X_i$, also called *belief updating*. For this query, $f(X)$ equals a $\delta$-function, and the above equation reduces to counting the fraction of occurrences of $X_i = x_i$ in the samples,

$$\hat{P}(x_i|e) = \frac{1}{T} \sum_{t=1}^{T} \delta(x_i|x^{(t)}) \ , \tag{2}$$

where $\delta(x_i|x^{(t)})=1$ iff $x_i = x_i^{(t)}$ and $\delta(x_i|x^{(t)})=0$ otherwise. Alternatively, a *mixture estimator* can be used,

$$\hat{P}(x_i|e)] = \frac{1}{T} \sum_{t=1}^{T} P(x_i|x_{-i}^{(t)}) \ , \tag{3}$$

where $x_{-i}^{(t)} = x^{(t)} \backslash x_i$.

A significant limitation of sampling, however, is that the statistical variance increases when the number of variables in the network grows and therefore the number of samples necessary for accurate estimation increases. In this paper, we present a sampling scheme for Bayesian networks with discrete variables that reduces the sampling variance by sampling from a subset of the variables, a technique also known as *collapsing* or *Rao-Blackwellisation*. The fundamentals of Rao-Blackwellised sampling were developed by Casella and Robert (1996) and Liu, Wong, and Kong (1994) for Gibbs sampling and MacEachern, Clyde, and Liu (1998) and Doucet, Gordon, and Krishnamurthy (1999) for importance sampling. Doucet, de Freitas, Murphy, and Russell (2000) extended Rao-Blackwellisation to Particle Filtering in Dynamic Bayesian networks.

The basic Rao-Blackwellisation scheme can be described as follows. Suppose we partition the space of variables $X$ into two subsets $C$ and $Z$. Subsequently, we can re-write any function $f(X)$ as $f(C, Z)$. If we can generate samples from distribution $P(C|e)$ and compute $E[f(C, Z)|c, e]$, then we can perform sampling on subset $C$ only, generating samples $c^{(1)}, c^{(2)}, ..., c^{(T)}$ and approximating the quantity of interest by

$$E[f(C, Z)|e] = E_C[E_Z[f(C, Z)|c, e]] \approx \hat{f}(X) = \frac{1}{T} \sum_{t=1}^{T} E_Z[f(C, Z)|c^{(t)}, e] \ . \tag{4}$$

The posterior marginals' estimates for the cutset variables can be obtained using an expression similar to Eq.(2),

$$\hat{P}(c_i|e) = \frac{1}{T} \sum_t \delta(c_i|c^{(t)}) \ , \tag{5}$$

or using a mixture estimator similar to Eq.(3),

$$\hat{P}(c_i|e) = \frac{1}{T} \sum_t P(c_i|c_{-i}^{(t)}, e) \ . \tag{6}$$

For $X_i \in X \backslash C, E, E[P(X_i|e)] = E_C[P(X_i|c, e)]$ and Eq.(4) becomes

$$\hat{P}(X_i|e) = \frac{1}{T} \sum_t P(X_i|c^{(t)}, e) \ . \tag{7}$$

Since the convergence rate of Gibbs sampler is tied to the maximum correlation between two samples (Liu, 2001), we can expect an improvement in the convergence rate when sampling in a lower dimensional space since 1) some of the highly-correlated variables may be marginalized out and 2) the dependencies between the variables inside a smaller set are likely to be weaker because the variables will be farther apart and their sampling distributions will be smoothed out. Additionally, the estimates obtained from sampling in a lower dimensional space can be expected to have lower sampling variance and therefore require fewer samples to achieve the same accuracy of the estimates. On the other hand, the cost of generating each sample may increase. Indeed, the principles of Rao-Blackwellised sampling have been applied only in a few classes of probabilistic models with specialized structure (Kong, Liu, & Wong, 1994; Escobar, 1994; MacEachern, 1994; Liu, 1996; Doucet & Andrieu, 2001; Andrieu, de Freitas, & Doucet, 2002; Rosti & Gales, 2004).

The contribution of this paper is in presenting a general, structure-based scheme which applies the Rao-Blackwellisation principle to Bayesian networks. The idea is to exploit the property that conditioning on a subset of variables simplifies the network's structure, allowing efficient query processing by exact algorithms. In general, exact inference by variable elimination (Dechter, 1999a, 2003) or join-tree algorithms (Lauritzen & Spiegelhalter, 1988; Jensen, Lauritzen, & Olesen, 1990) is time and space exponential in the induced-width $w$ of the network. However, when a subset of the variables is assigned (i.e., conditioned upon) the induced-width of the conditioned network may be reduced.

The idea of cutset sampling is to choose a subset of variables $C$ such that conditioning on $C$ yields a sparse enough Bayesian network having a small induced width to allow exact inference. Since a sample is an assignment to all cutset variables, we can efficiently generate a new sample over the cutset variables in the conditioned network where the computation of $P(c|e)$ and $P(X_i|c, e)$ can be bounded. In particular, if the sampling set $C$ cuts all the cycles in the network (i.e., it is a loop-cutset), inference over the conditioned network becomes linear. In general, if $C$ is a $w$-cutset, namely a subset of nodes such that when assigned, the induced-width of the conditioned network is $w$, the time and space complexity of computing the next sample is $O(|C| \cdot N \cdot d^{w+2})$ where $d$ is the maximum domain size and $N = |X|$.

The idea of exploiting properties of conditioning on a subset of variables was first proposed for exact belief updating in the context of cutset-conditioning (Pearl, 1988). This scheme requires enumerating all instantiations of the cutset variables. Since the number of instances is exponential in the size of the cutset $|C|$, sampling over the cutset space may be the right compromise when the size of the cutset is too big. Thus, sampling on a cutset can also be viewed as an anytime approximation of the cutset-conditioning approach.

Although Rao-Blackwellisation in general and cutset sampling in particular can be applied in the context of any sampling algorithm, we will introduce the principle in the context of Gibbs sampling (Geman & Geman, 1984; Gilks, Richardson, & Spiegelhalter, 1996; MacKay, 1996), a Markov Chain Monte Carlo sampling method for Bayesian networks. Extension to any other sampling approach or any other graphical models, such as Markov networks, should be straight forward. We recently demonstrated how the idea can be incorporated into importance sampling (Bidyuk & Dechter, 2006).

The paper defines and analyzes the cutset sampling scheme and investigates empirically the trade-offs between sampling and exact computation over a variety of randomly generated networks and grid structure networks as well as known real-life benchmarks such as CPCS networks and coding networks. We show that cutset sampling converges faster than pure sampling in terms of the number of samples, as dictated by theory, and is also almost always time-wise cost effective on all the benchmarks tried. We also demonstrate the applicability of this scheme to some deterministic networks, such as Hailfinder network and coding networks, where the Markov Chain is non-ergodic and Gibbs sampling does not converge.

Section 2 provides background information. Specifically, section 2.1 introduces Bayesian networks, section 2.2 reviews exact inference algorithms for Bayesian networks, and section 2.3 provides background on Gibbs sampling. The contribution of the paper presenting the cutset sampling starts in section 3. Section 4 presents the empirical evaluation of cutset sampling. We also present an empirical evaluation of the sampling variance and the resulting standard error based on the method of *batch means* (for more details, see Geyer, 1992).

In section 5, we review previous application of Rao-Blackwellisation and section 6 provides summary and conclusions.

## 2. Background

In this section, we define essential terminology and provide background information on Bayesian networks.

### 2.1 Preliminaries

We use upper case letters without subscripts, such as $X$, to denote sets of variables and lower case letters without subscripts to denote an instantiation of a group of variables (e.g., $x$ indicates that each variable in set $X$ is assigned a value). We use an upper case letter with a subscript, such as $X_i$, to denote a single variable and a lower case letter with a subscript, such as $x_i$, to denote an instantiated variable (e.g., $x_i$ denotes an arbitrary value in the domain of $X_i$ and means $X_i = x_i$). $\mathcal{D}(X_i)$ denotes the domain of the variable $X_i$. A superscript in a subscripted lower case letter would be used to distinguish different specific values for a variable, i.e., $\mathcal{D}(X_i) = \{x_i^1, x_i^2, ...\}$. We will use $x$ to denote an instantiation of a set of variables $x = \{x_1, ..., x_{i-1}, x_i, x_{i+1}, ..., x_n\}$ and $x_{-i} = x \backslash x_i$ to denote $x$ with element $x_i$ removed. Namely, $x_{-i} = \{x_1, x_2, ..., x_{i-1}, x_{i+1}, ..., x_n\}$.

DEFINITION **2.1 (graph concepts)** *A* **directed graph** *is a pair* $D = <V, E>$, *where* $V = \{X_1, ..., X_n\}$ *is a set of nodes and* $E = \{(X_i, X_j) | X_i, X_j \in V\}$ *is the set of edges. Given* $(X_i, X_j) \in E$, $X_i$ *is called a* **parent** *of* $X_j$, *and* $X_j$ *is called a* **child** *of* $X_i$. *The set of* $X_i$'s *parents is denoted* $pa(X_i)$, *or* $pa_i$, *while the set of* $X_i$'s *children is denoted* $ch(X_i)$, *or* $ch_i$. *The family of* $X_i$ *includes* $X_i$ *and its parents. The* **moral graph** *of a directed graph* $D$ *is the undirected graph obtained by connecting the parents of all the nodes in* $D$ *and removing the arrows. A* **cycle-cutset** *of an undirected graph a subset of nodes that, when removed, yields a graph without cycles. A* **loop** *in a directed graph* $D$ *is a subgraph of* $D$ *whose underlying graph is a cycle. A directed graph is acyclic if it has no directed loops. A directed graph is* **singly-connected** *(also called a* **poly-tree**), *if its underlying undirected graph has no cycles. Otherwise, it is called* **multiply-connected**.

DEFINITION **2.2 (loop-cutset)** *A vertex* $v$ *is a* **sink** *with respect to a loop* $\mathcal{L}$ *if the two edges adjacent to* $v$ *in* $\mathcal{L}$ *are directed into* $v$. *A vertex that is not a sink with respect to a loop* $\mathcal{L}$ *is called an* allowed *vertex with respect to* $\mathcal{L}$. *A* **loop-cutset** *of a directed graph* $D$ *is a set of vertices that contains at least one allowed vertex with respect to each loop in* $D$.

DEFINITION **2.3 (Belief Networks)** *Let* $X = \{X_1, ..., X_n\}$ *be a set of random variables over multi-valued domains* $\mathcal{D}(X_1), ..., \mathcal{D}(X_n)$. *A* **belief network** *(BN) (Pearl, 1988) is a pair* $<G, P>$ *where* $G$ *is a directed acyclic graph whose nodes are the variables* $X$ *and* $P = \{P(X_i | pa_i) | i = 1, ..., n\}$ *is the set of conditional probability tables (CPTs) associated with each* $X_i$. *The BN represents a joint probability distribution having the product form:*

$$P(x_1, ...., x_n) = \prod_{i=1}^{n} P(x_i | pa(X_i))$$

*An evidence* $e$ *is an instantiated subset of variables* $E \subset X$.

The structure of the directed acyclic graph $G$ reflects the dependencies between the variables using $d$-separation criterion. The parents of a variable $X_i$ together with its children and parents of its children form a **Markov blanket**, denoted $markov_i$, of node $X_i$. We will use $x_{markov_i}$ to denote $x$ restricted to variables in $markov_i$. We know that the node $X_i$ is independent of the rest of the variables conditioned on its Markov blanket. Namely, $P(x_i|x_{-i}) = P(x_i|x_{markov_i})$.

The most common query over belief networks is *belief updating* which is the task of computing the posterior distribution $P(X_i|e)$ given evidence $e$ and a query variable $X_i \in X$. Reasoning in Bayesian networks is NP-hard (Cooper, 1990). Finding approximate posterior marginals with a fixed accuracy is also NP-hard (Dagum & Luby, 1993; Abdelbar & Hedetniemi, 1998). When the network is a poly-tree, belief updating and other inference tasks can be accomplished in time linear in size of the input. In general, exact inference is exponential in the induced width of the network's moral graph.

DEFINITION **2.4 (induced-width)** *The* **width of a node** *in an ordered undirected graph is the number of the node's neighbors that precede it in the ordering. The* **width of an ordering** $d$, *denoted* $w(d)$, *is the maximum width over all nodes. The* **induced width** *of an ordered graph,* $w^*(d)$, *is the width of the ordered graph obtained by processing the nodes from last to first as follows: when node $X$ is processed, all its preceding neighbors are connected. The resulting graph is called* **induced graph** *or* **triangulated graph**.

The task of finding the minimal induced width of a graph (over all possible orderings) is NP-complete (Arnborg, 1985).

## 2.2 Reasoning in Bayesian Networks

Belief propagation algorithm, which we introduce in Section 2.2.2 below, performs belief updating in singly-connected Bayesian networks in time linear in the size of the input (Pearl, 1988). In loopy networks, the two main approaches for belief updating are *cutset conditioning* and *tree clustering*. These algorithms are often referred to as "inference" algorithms. We will briefly describe the idea of clustering algorithms in Section 2.2.1 and the conditioning method in Section 2.2.3.

### 2.2.1 Variable Elimination and Join-Tree Clustering (JTC)

The join-tree clustering approach (JTC) refers to a family of algorithms including join-tree propagation (Lauritzen & Spiegelhalter, 1988; Jensen et al., 1990) and bucket-tree elimination (Dechter, 2003, 1999a). The idea is to first obtain a tree-decomposition of the network into clusters of functions connected as a tree and then propagate messages between the clusters in the tree. The tree-decomposition is a singly-connected undirected graph whose nodes, also called clusters, contain subsets of variables and input functions defined over those variables. The tree-decomposition must contain each function once and satisfy running intersection property (Maier, 1983). For a unifying perspective of tree-decomposition schemes see (Zhang & Poole, 1994; Dechter, 1999b; Kask, Dechter, Larrosa, & Dechter, 2005).

Given a tree-decomposition of the network, the message propagation over this tree can be synchronized. We select any one cluster as the root of the tree and propagate messages up and down the tree. A message from cluster $V_i$ to neighbor $V_j$ is a function over the separator set $V_i \cap V_j$ that is a marginalization of the product of all functions in $V_i$ and all messages that $V_i$ received from its neighbors besides $V_j$. Assuming that the maximum number of variables in a cluster is $w + 1$ and maximum domain size is $d$, the time and space required to process one cluster is $O(d^{(w+1)})$. Since the maximum number of clusters is bounded by $|X| = N$, the complexity of variable-elimination algorithms and cluster-tree propagation schemes is $O(N \cdot d^{(w+1)})$. The parameter $w$, the maximum cluster size minus 1, is called the tree-width of the tree decomposition. The minimal tree-width is identical to the minimal induced width of a graph.

### 2.2.2 ITERATIVE BELIEF PROPAGATION (IBP)

Belief propagation (BP) is an iterative message-passing algorithm that performs exact inference for singly-connected Bayesian networks (Pearl, 1988). In each iteration, every node $X_i$ sends a $\pi_j(X_i)$ message to each child $j$ and receives a $\lambda_j(X_i)$ message from each child. The message-passing order can be organized so that it converges in two iterations. In essence the algorithm is the same as the join-tree clustering approach applied directly to the poly-tree. Applied to Bayesian networks with loops, the algorithm usually iterates longer (until it may converge) and hence, is known as Iterative Belief Propagation (IBP) or loopy belief propagation. IBP provides no guarantees on convergence or quality of approximate posterior marginals but was shown to perform well in practice (Rish, Kask, & Dechter, 1998; Murphy, Weiss, & Jordan, 1999). It is considered the best algorithm for inference in coding networks (Frey & MacKay, 1997; Kschischang & Frey, 1998) where finding the most probable variable values equals the decoding process (McEliece, MacKay, & Cheng, 1997). Algorithm IBP requires linear space and usually converges fast if it converges. In our benchmarks, IBP converged within 25 iterations or less (see Section 4).

### 2.2.3 CUTSET CONDITIONING

When the tree-width $w$ of the Bayesian network is too large and the requirements of inference schemes such as bucket elimination and join-tree clustering (JTC) exceed available memory, we can switch to the alternative *cutset conditioning* schemes (Pearl, 1988; Peot & Shachter, 1992; Shachter, Andersen, & Solovitz, 1994). The idea of cutset conditioning is to select a subset of variables $C \subset X \backslash E$, the cutset, and obtain posterior marginals for any node $X_i \in X \backslash C, E$ using:

$$P(x_i|e) = \sum_{c \in \mathcal{D}(C)} P(x_i|c, e)P(c|e) \qquad (8)$$

Eq.(8) above implies that we can enumerate all instantiations over $C$, perform exact inference for each cutset instantiation $c$ to obtain $P(x_i|c, e)$ and $P(c|e)$ and then sum up the results. The total computation time is exponential in the size of the cutset because we have to enumerate all instantiations of the cutset variables.

If $C$ is a loop-cutset, then, when the nodes in $C$ are assigned, the Bayesian network can be transformed into an equivalent poly-tree and $P(x_i|c, e)$ and $P(c|e)$ can be computed via BP in time and space linear in the size of the network. For example, the subset $\{A, D\}$ is a loop-cutset of the belief network shown in Figure 1, left, with evidence $E = e$. On the right,

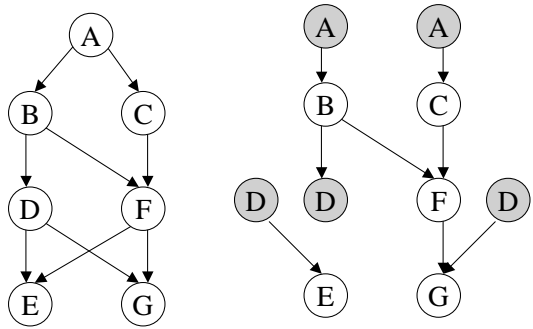

Figure 1: When nodes $A$ and $D$ in the loopy Bayesian network (left) are instantiated, the network can be transformed into an equivalent singly-connected network (right). In the transformation process, a replica of an observed node is created for each child node.

Figure 1 shows an equivalent singly-connected network resulting from assigning values to $A$ and $D$.

It is well-known that the minimum induced width $w^*$ of the network is always less than the size of the smallest loop-cutset (Bertele & Brioschi, 1972; Dechter, 2003). Namely, $w^* + 1 \leq |C|$ for any $C$. Thus, inference approaches (e.g., bucket elimination) are never worse and often are better than cutset conditioning time-wise. However, when $w^*$ is too large we must resort to cutset conditioning search in order to trade space for time. Those considerations yield a hybrid search and inference approach. Since observed variables can break down the dependencies in the network, a network with an observed subset of variables $C$ often can be transformed into an equivalent network with a smaller induced width, $w_C$, which we will term the *adjusted induced width*. Hence, when any subset of variables $C \subset X$ is observed, complexity is bounded exponentially by the adjusted induced width of the graph $w_C$.

Definition 2.5 (adjusted induced width) *Given a graph $G=<X,E>$, the* **adjusted induced width** *of $G$ relative to $C$, denoted $w_C$, is its induced width once $C$ is removed from its moral graph.*

Definition 2.6 ($w$-cutset) *Given a graph $G=<X,E>$, a subset of nodes $C \subset X$ is a* $w$**-cutset** *of $G$ if its adjusted induced width equals $w$.*

If $C$ is a $w$-cutset, the quantities $P(x_i|c,e)$ and $P(c|e)$ can be computed in time and space exponential in $w$, which can be much smaller than the tree-width of the unconditioned network. The resulting scheme requires memory exponential in $w$ and time $O(d^{|C|} \cdot N \cdot d^{(w+1)})$ where $N$ is the size of the network and $d$ is the maximum domain size. Thus, the performance can be tuned to the available system memory resource via the bounding parameter $w$.

Given a constant $w$, finding a minimal $w$-cutset (to minimize the cutset conditioning time) is also a hard problem. Several greedy heuristic approaches have been proposed by Geiger and Fishelson (2003) and by Bidyuk and Dechter (2003, 2004). We elaborate more in Section 3.5.

## 2.3 Gibbs Sampling

Since the complexity of inference algorithms is memory exponential in the network's induced width (or tree-width) and since resorting to the cutset-conditioning scheme may take too much time when the $w$-cutset size is too large, we must often resort to approximation methods. Sampling methods for Bayesian networks are commonly used approximation techniques. This section provides background on Gibbs sampling, a Markov Chain Monte Carlo method, which is one of the most popular sampling schemes and is the focus of this paper. Although the method may be applied to the networks with continuous distributions, we limit our attention in this paper to discrete random variables with finite domains.

### 2.3.1 GIBBS SAMPLING FOR BAYESIAN NETWORKS

---

*Ordered Gibbs Sampler*
**Input:** A belief network $\mathcal{B}$ over $X=\{X_1,...,X_n\}$ and evidence $e=\{(X_i = e_i)|X_i \in E \subseteq X\}$.
**Output:** A set of samples $\{x^{(t)}\}$, $t = 1...T$.
1. **Initialize:** Assign random value $x_i^{(0)}$ to each variable $X_i \in X \backslash E$ from $\mathcal{D}(X_i)$. Assign evidence variables their observed values.
2. **Generate samples:**
For t = 1 to T, generate a new sample $x^{(t)}$:
    For i = 1 to N, compute a new value $x_i^{(t)}$ for variable $X_i$:
        Compute distribution $P(X_i|x_{markov_i}^{(t)})$ and sample $x_i^{(t)} \leftarrow P(X_i|x_{markov_i}^{(t)})$.
        Set $X_i = x_i^{(t)}$.
    End For $i$
End For $t$

---

Figure 2: A *Gibbs sampling* Algorithm

Given a Bayesian network over the variables $X = \{X_1,...,X_n\}$, and evidence $e$, Gibbs sampling (Geman & Geman, 1984; Gilks et al., 1996; MacKay, 1996) generates a set of samples $\{x^{(t)}\}$ where each sample $x^{(t)} = \{x_1^{(t)},...,x_n^{(t)}\}$ is an instantiation of all the variables. The superscript $t$ denotes a sample index and $x_i^{(t)}$ is the value of $X_i$ in sample $t$. The first sample can be initialized at random. When generating a new sample from sample $x_i^{(t)}$, a new value for variable $X_i$ is sampled from probability distribution $P(X_i|x_{-i}^{(t)})$ (recall that $P(X_i|x_{-i}^{(t)}) = P(X_i|x_1^{(t+1)},...,x_{i-1}^{(t+1)},x_{i+1}^{(t)},...,x_n^{(t)})$) which we will denote as $x_i \leftarrow P(X_i|x_{-i}^{(t)})$. The next sample $x_i^{(t+1)}$ is generated from the previous sample $x_i^{(t)}$ following one of two schemes.

**Random Scan Gibbs Sampling.** Given a sample $x^{(t)}$ at iteration $t$, pick a variable $X_i$ at random and sample a new value $x_i$ from the conditional distribution $x_i \leftarrow P(X_i|x_{-i}^{(t)})$ leaving other variables unchanged.

**Systematic Scan (Ordered) Gibbs Sampling.** Given a sample $x^{(t)}$, sample a new value for each variable in some order:

$$x_1 \quad \leftarrow \quad P(X_1|x_2^{(t)}, x_3^{(t)}, ..., x_n^{(t)})$$

$$x_2 \quad \leftarrow \quad P(X_2|x_1^{(t+1)}, x_3^{(t)}, ..., x_n^{(t)})$$

$$...$$

$$x_i \quad \leftarrow \quad P(X_i|x_1^{(t+1)}, ..., x_{i-1}^{(t+1)}, x_{i+1}^{(t)}, ..., x_n^{(t)})$$

$$...$$

$$x_n \quad \leftarrow \quad P(X_n|x_1^{(t+1)}, x_2^{(t+1)}, ..., x_{n-1}^{(t+1)})$$

In Bayesian networks, the conditional distribution $P(X_i|x_{-i}^{(t)})$ is dependent only on the assignment to the Markov blanket of variable $X_i$. Thus, $P(X_i|x_{-i}^{(t)}) = P(X_i|x_{markov_i}^{(t)})$ where $x_{markov_i}^{(t)}$ is the restriction of $x^{(t)}$ to $markov_i$. Given a Markov blanket of $X_i$, the sampling probability distribution is given explicitly by Pearl (1988):

$$P(x_i|x_{markov_i}^{(t)}) = \alpha P(x_i|x_{pa_i}^{(t)}) \prod_{\{j|X_i \in ch_j\}} P(x_j^{(t)}|x_{pa_j}^{(t)}) \qquad (9)$$

Thus, generating a complete new sample can be done in $O(n \cdot r)$ multiplication steps where $r$ is the maximum family size and $n$ is the number of variables.

The sequence of samples $x^{(1)}, x^{(2)}, ...$ can be viewed as a sequence of states in a Markov chain. The transition probability from state $\{x_1^{(t+1)}, ..., x_{i-1}^{(t+1)}, x_i^{(t)}, x_{i+1}^{(t)}, ..., x_n^{(t)}\}$ to state $\{x_1^{(t+1)}, ..., x_{i-1}^{(t+1)}, x_i^{(t+1)}, x_{i+1}^{(t)}, ..., x_n^{(t)}\}$ is defined by the sampling distribution $P(X_i|x_{-i}^{(t)})$. By construction, a Markov chain induced by Gibbs sampling has an invariant distribution $P(X|e)$. However, since the values assigned by the Gibbs sampler to variables in a sample $x^{(t+1)}$ depend on the assignment of values in the previous sample $x^{(t)}$, it follows that the sample $x^{(n)}$ depends on the initial state $x^{(0)}$. The convergence of the Markov chain is defined by the rate at which the distance between the distribution $P(x^{(n)}|x^{(0)}, e)$ and the stationary distribution $P(X|e)$ (i.e., variational distance, $L^1$-distance, or $\chi^2$) converges to 0 as a function of $n$. Intuitively, it reflects how quickly the inital state $x^{(0)}$ can be "forgotten." The convergence is guaranteed as $T \rightarrow \infty$ if Markov chain is *ergodic* (Pearl, 1988; Gelfand & Smith, 1990; MacKay, 1996). A Markov chain with a finite number of states is *ergodic* if it is *aperiodic* and *irreducible* (Liu, 2001). A Markov chain is aperiodic if it does not have regular loops. A Markov chain is irreducible if we can get from any state $S_i$ to any state $S_j$ (including $S_i$) with non-zero probability in a finite number of steps. The irreducibility guarantees that we will be able to visit (as number of samples increases) all statistically important regions of the state space. In Bayesian networks, the conditions are almost always satisfied as long as all conditional probabilities are positive (Tierney, 1994).

To ensure that the collected samples are drawn from distribution close to $P(X|e)$, a "burn-in" time may be allocated. Namely, assuming that it takes $\approx K$ samples for the Markov chain to get close to the stationary distribution, the first $K$ samples may not be included into the computation of posterior marginals. However, determining $K$ is hard (Jones & Hobert, 2001). In general, the "burn-in" is optional in the sense that the convergence of the estimates to the correct posterior marginals does not depend on it. For completeness sake, the algorithm is given in Figure 2.

When convergence conditions are satisfied, an ergodic average $f_T(X) = \frac{1}{T} \sum_t f(x^t)$ for any function $f(X)$ is guaranteed to converge to the expected value $E[f(X)]$ as $T$ increases.

In other words, $|f_T(X) - E[f(X)]| \to 0$ as $T \to \infty$. For a finite-state Markov chain that is irreducible and aperiodic, the following result applies (see Liu, 2001, Theorem 12.7.2):

$$\sqrt{T}|f_T(X) - E[f(X)]| \to N(0, \sigma(f)^2) \tag{10}$$

for any initial assignment to $x^{(0)}$. The variance term $\sigma(f)^2$ is defined as follows:

$$\sigma(f)^2 = 2\tau(f)\sigma^2$$

where $\sigma^2 = var[f(X)]$ and $\tau(h)$ is the integrated autocorrelation time.

Our focus is on computing the posterior marginals $P(X_i|e)$ for each $X_i \in X \backslash E$. The posterior marginals can be estimated using either a *histogram estimator*:

$$\hat{P}(X_i = x_i|e) = \frac{1}{T}\sum_{t=1}^{T}\delta(x_i|x^{(t)}) \tag{11}$$

or a *mixture estimator*:

$$\hat{P}(X_i = x_i|e) = \frac{1}{T}\sum_{t=1}^{T}P(x_i|x_{-i}^{(t)}) \tag{12}$$

The histogram estimator corresponds to counting samples where $X_i = x_i$, namely $\delta(x_i|x^{(t)}) = 1$ if $x_i^{(t)} = x_i$ and equals 0 otherwise. Gelfand and Smith (1990) have pointed out that since mixture estimator is based on estimating conditional expectation, its sampling variance is smaller due to Rao-Blackwell theorem. Thus, mixture estimator should be preferred. Since $P(x_i|x_{-i}^{(t)}) = P(x_i|x_{markov_i}^{(t)})$, the mixture estimator is simply an average of conditional probabilities:

$$\hat{P}(x_i|e) = \frac{1}{T}\sum_{t=1}^{T}P(x_i|x_{markov_i}^{(t)}) \tag{13}$$

As mentioned above, when the Markov chain is ergodic, $\hat{P}(X_i|e)$ will converge to the exact posterior marginal $P(X_i|e)$ as the number of samples increases. It was shown by Roberts and Sahu (1997) that random scan Gibbs sampler can be expected to converge faster than the systematic scan Gibbs sampler. Ultimately, the convergence rate of Gibbs sampler depends on the correlation between two consecutive samples (Liu, 1991; Schervish & Carlin, 1992; Liu et al., 1994). We review this subject in the next section.

## 2.4 Variance Reduction Schemes

The convergence rate of the Gibbs sampler depends on the strength of the correlations between the samples (which are also the states of the Markov chain). The term correlation is used here to mean that the samples are dependent, as mentioned earlier. In the case of a finite-state irreducible and aperiodic Markov chain, the convergence rate can be expressed through *maximal correlation* between states $x^{(0)}$ and $x^{(n)}$ (see Liu, 2001, ch. 12). In practice, the convergence rate can be analyzed through covariance $cov[f(x^{(t)}), f(x^{(t+1)})]$, where $f$ is some function, also called auto-covariance.

The convergence of the estimates to the exact values depends on both the convergence rate of the Markov chain to the stationary distribution and the variance of the estimator.

Both of these factors contribute to the value of the term $\sigma(f)^2$ in Eq.(10). The two main approaches that allow to reduce correlation between samples and reduce sampling variance of the estimates are *blocking* (grouping variables together and sampling simultaneously) and *collapsing* (integrating out some of the random variables and sampling a subset), also known as *Rao-Blackwellisation*.

Given a joint probability distribution over three random variables $X$, $Y$, and $Z$, we can depict the essence of those three sampling schemes as follows:

1. Standard Gibbs:

$$x^{(t+1)} \quad \leftarrow \quad P(X|y^{(t)}, z^{(t)}) \tag{14}$$
$$y^{(t+1)} \quad \leftarrow \quad P(Y|x^{(t+1)}, z^{(t)}) \tag{15}$$
$$z^{(t+1)} \quad \leftarrow \quad P(Z|x^{(t+1)}, y^{(t+1)}) \tag{16}$$

2. Collapsed (variable $Z$ is integrated out):

$$x^{(t+1)} \quad \leftarrow \quad P(X|y^{(t)}) \tag{17}$$
$$y^{(t+1)} \quad \leftarrow \quad P(Y|x^{(t+1)}) \tag{18}$$

3. Blocking by grouping $X$ and $Y$ together:

$$(x^{(t+1)}, y^{(t+1)}) \quad \leftarrow \quad P(X, Y|z^{(t)}) \tag{19}$$
$$z^{(t+1)} \quad \leftarrow \quad P(Z|x^{(t+1)}, y^{(t+1)}) \tag{20}$$

The blocking reduces the correlation between samples by grouping highly correlated variables into "blocks." In collapsing, the highly correlated variables are marginalized out, which also results in the smoothing of the sampling distributions of the remaining variables ($P(Y|x)$ is smoother than $P(Y|x, z)$). Both approaches lead to reduction of the sampling variance of the estimates, speeding up their convergence to the exact values.

Generally, blocking Gibbs sampling is expected to converge faster than standard Gibbs sampler (Liu et al., 1994; Roberts & Sahu, 1997). Variations on this scheme have been investigated by Jensen et al. (1995) and Kjaerulff (1995). Given the same number of samples, the estimate resulting from collapsed Gibbs sampler is expected to have lower variance (converge faster) than the estimate obtained from blocking Gibbs sampler (Liu et al., 1994). Thus, collapsing is preferred to blocking. The analysis of the collapsed Gibbs sampler can be found in Escobar (1994), MacEachern (1994), and Liu (1994, 1996).

The caveat in the utilization of the collapsed Gibbs sampler is that computation of the probabilities $P(X|y)$ and $P(Y|x)$ must be efficient time-wise. In case of Bayesian networks, the task of integrating out some variables is equivalent to posterior belief updating where evidence variables and sampling variables are observed. Its time complexity is therefore exponential in the adjusted induced width, namely, in the effective width of the network after some dependencies are broken by instantiated variables (evidence and sampled).

## 2.5 Importance Sampling

Since sampling from the target distribution is hard, different sampling methods explore different trade-offs in generating samples and obtaining estimates. As we already discussed, Gibbs sampling generates dependent samples but guarantees convergence of the sampling distribution to the target distribution. Alternative approach, called importance sampling, is to generate samples from a sampling distribution $Q(X)$ that is different from $P(X|e)$ and include the weight $w^{(t)} = P(x^{(t)}|e)/Q(x^{(t)})$ of each sample $x^{(t)}$ in the computation of the estimates as follows:

$$\hat{f}_T(X) = \frac{1}{T} \sum_{t=1}^{T} f(x^t) w^{(t)} \tag{21}$$

The convergence of $\hat{f}_T(X)$ to $E[f(X)]$ is guaranteed as long as the condition $P(x|e) \neq 0 \Rightarrow Q(x) \neq 0$ holds. The convergence speed depends on the distance between $Q(X)$ and $P(X|e)$.

One of the simplest forms of importance sampling for Bayesian networks is likelihood weighting (Fung & Chang, 1989; Shachter & Peot, 1989) which processes variables in topological order, sampling root variables from their priors and the remaining variables from conditional distribution $P(X_i|pa_i)$ defined by their conditional probability table (the evidence variables are assigned their observed values). Its sampling distribution is close to the prior and, as a result, it usually converges slowly when the evidence is concentrated around the leaf nodes (nodes without children) and when the probability of evidence is small. Adaptive (also called dynamic) importance sampling is a method that attempts to speed up the convergence by updating the sampling distribution based on the weight of previously generated samples. Adaptive importance sampling methods include self-importance sampling, heuristic importance sampling (Shachter & Peot, 1989), and, more recently, AIS-BN (Cheng & Druzdzel, 2000) and EPIS-BN (Yuan & Druzdzel, 2003). In the empirical section, we compare the performance of the proposed cutset sampling algorithm with AIS-BN which is considered a state-of-the-art importance sampling algorithm to date (although EPIS-BN was shown to perform better in some networks) and, hence, describe AIS-BN here in more detail.

AIS-BN algorithm is based on the observation that if we could sample each node in topological order from distribution $P(X_i|pa_i, e)$, then the resulting sample would be drawn from the target distribution $P(X|e)$. Since this distribution is unknown for any variable that has observed descendants, AIS-BN initializes the sampling distributions $P^0(X_i|pa_i, e)$ equal to either $P(X_i|pa_i)$ or a uniform distribution and then updates each distribution $P^k(X_i|pa_i, e)$ every $l$ samples so that the next sampling distribution $P^{k+1}(X_i|pa_i, e)$ will be closer to $P(X_i|pa_i, e)$ than $P^k(X_i|pa_i, e)$ as follows:

$$P^{k+1}(x_i|pa_i, e) = P^k(x_i|pa_i, e) + \eta(k) \cdot (P'(x_i|pa_i, e) - P^k(x_i|pa_i, e))$$

where $\eta(k)$ is a positive function that determines the learning rate and $P'(x_i|pa_i, e)$ is an estimate of $P(x_i|pa_i, e)$ based on the last $l$ samples.

## 3. Cutset Sampling

This section presents the cutset sampling scheme. As we discussed above, sampling on a cutset is guaranteed to be more statistically efficient. Cutset sampling scheme is a computationally efficient way of sampling from a "collapsed" variable subset $C \subset X$, tying the complexity of sample generation to the structure of the Bayesian network.

### 3.1 Cutset Sampling Algorithm

The cutset sampling scheme partitions the variable set $X$ into two subsets $C$ and $X \backslash C$. The objective is to generate samples from space $C=\{C_1, C_2, ..., C_m\}$ where each sample $c^{(t)}$ is an instantiation of all the variables in $C$. Following the Gibbs sampling principles, we wish to generate a new sample $c^{(t)}$ by sampling a value $c_i^{(t)}$ from the probability distribution $P(C_i|c_{-i}^{(t)}) = P(C_i|c_1^{(t+1)}, c_2^{(t+1)}, ..., c_{i-1}^{(t+1)}, c_{i+1}^{(t)}, ..., c_m^{(t)})$. We will use left arrow to denote that value $c_i$ is drawn from distribution $P(C_i|c_{-i}^{(t)})$:

$$c_i \leftarrow P(C_i|c_{-i}^{(t)}, e) \tag{22}$$

If we can compute the probability distribution $P(C_i|c_{-i}^{(t)}, e)$ efficiently for each sampling variable $C_i \in C$, then we can generate samples efficiently. The relevant conditional distributions can be computed by exact inference whose complexity is tied to the network structure. We denote by $JTC(\mathcal{B}, X_i, e)$ a generic algorithm in the class of variable-elimination or join-tree clustering algorithms which, given a belief network $\mathcal{B}$ and evidence $e$, outputs the posterior probabilities $P(X_i|e)$ for variable $X_i \in X$ (Lauritzen & Spiegelhalter, 1988; Jensen et al., 1990; Dechter, 1999a). When the network's identity is clear, we will use the notation $JTC(X_i, e)$.

---

*Cutset Sampling*

**Input:** A belief network $\mathcal{B}$, a cutset $C = \{C_1, ..., C_m\}$, evidence e.

**Output:** A set of samples $c^t$, $t = 1...T$.

1. **Initialize:** Assign random value $c_i^0$ to each $C_i \in C$ and assign e.

2. **Generate samples:**

   For t = 0 to T-1, generate a new sample $c^{(t+1)}$ as follows:

   For i = 1 to m, compute new value $c_i^{(t)}$ for variable $C_i$ as follows:

   a. Compute $JTC(C_i, c_{-i}^{(t)}, e)$.

   b. Compute $P(C_i|c_{-i}^{(t)}, e) = \alpha P(C_i, c_{-i}^{(t)}, e)$.

   c. Sample:
   $$c_i^{(t+1)} \leftarrow P(C_i|c_{-i}^{(t)}, e) \tag{23}$$

   End For $i$

   End For $t$

---

Figure 3: *w-Cutset sampling* Algorithm

Therefore, for each sampling variable $C_i$ and for each value $c_i \in \mathcal{D}(C_i)$, we can compute $P(C_i, c_{-i}^{(t)}, e)$ via $JTC(C_i, c_{-i}^{(t)}, e)$ and obtain $P(C_i|c_{-i}^{(t)}, e)$ via normalization: $P(C_i|c_{-i}^{(t)}, e) = \alpha P(C_i, c_{-i}^{(t)}, e)$.

Cutset sampling algorithm that uses systematic scan Gibbs sampler is given in Figure 3. Clearly, it can be adapted to be used with the random scan Gibbs sampler as well. Steps (a)-(c) generate sample $(t+1)$ from sample $(t)$. For every variable $C_i \in C$ in sequence, the main computation is in step (a), where the distribution $P(C_i, c_{-i}^{(t)}, e)$ over $C_i$ is generated. This requires executing $JTC$ for every value $c_i \in \mathcal{D}(C_i)$, separately. In step (b), the conditional distribution is derived by normalization. Finally, step (c) samples a new value from the obtained distribution. Note that we only use $P(C_i|c_{-i}^{(t)}, e)$ as a short-hand notation for $P(C_i|c_1^{(t+1)}, ..., c_{i-1}^{(t+1)}, c_{i+1}^{(t)}, ..., c_k^{(t)}, e)$. Namely, when we sample a new value for variable $C_i$, the values of variables $C_1$ through $C_{i-1}$ have already been updated.

We will next demonstrate the process using the special case of loop-cutset (see Definition 2.1).

**Example 3.1** *Consider the belief network previously shown in Figure 1 with the observed node $E = e$ and loop-cutset $\{A, D\}$. We begin the sampling process by initializing sampling variables to $a^{(0)}$ and $d^{(0)}$. Next, we compute new sample values $a^{(1)}, d^{(1)}$ as follows:*

$$
\begin{align}
P(A|d^{(0)}, e) &= \alpha P_{JTC}(A, c^{(0)}, e) \tag{24} \\
a^{(1)} &\leftarrow P(A|d^{(0)}, e) \tag{25} \\
P(D|a^{(1)}, e) &= \alpha P_{JTC}(D, a^{(1)}, e) \tag{26} \\
d^{(1)} &\leftarrow P(D|a^{(1)}, e) \tag{27}
\end{align}
$$

*The process above corresponds to two iterations of the inner loop in Figure 3. Eq. (24)-(25), where we sample a new value for variable $A$, correspond to steps (a)-(c) of the first iteration. In the second iteration, Eq.(26)-(27), we sample a new value for variable $D$. Since the conditioned network is a poly-tree (Figure 1, right), computing probabilities $P_{JTC}(A|d^{(t)}, e)$ and $P_{JTC}(D|a^{(t+1)}, e)$ via JTC reduces to Pearl's belief propagation algorithm and the distributions can be computed in linear time.*

### 3.2 Estimating Posterior Marginals

Once a set of samples over a subset of variables $C$ is generated, we can estimate the posterior marginals of any variable in the network using mixture estimator. For sampling variables, the estimator takes the form similar to Eq.(12):

$$
\hat{P}(C_i|e) = \frac{1}{T} \sum_{t=1}^{T} P(C_i|c_{-i}^{(t)}, e) \tag{28}
$$

For variables in $X \backslash C, E$, the posterior marginal estimator is:

$$
\hat{P}(X_i|e) = \frac{1}{T} \sum_{t=1}^{T} P(X_i|c^{(t)}, e) \tag{29}
$$

We can use $JTC(X_i, c^{(t)}, e)$ to obtain the distribution $P(X_i|c^{(t)}, e)$ over the input Bayesian network conditioned on $c^{(t)}$ and $e$ as shown before.

If we maintain a running sum of the computed distributions $P(C_i|c_{-i}^{(t)}, e)$ and $P(X_i|c^{(t)}, e)$ during sample generation, the sums in the right hand side of Eq.(28)-(29) will be readily available. As we noted before, the estimators $\hat{P}(C_i|e)$ and $\hat{P}(X_i|e)$ are guaranteed to converge to their corresponding exact posterior marginals as $T$ increases as long as the Markov

chain over the cutset $C$ is ergodic. While for the cutset variables the estimator is a simple ergodic average, for $X_i \in X \backslash C, E$ the convergence can also be derived directly from first principles:

THEOREM **3.2** *Given a Bayesian network $\mathcal{B}$ over $X$, evidence variables $E \subset X$, and cutset $C \subset X \backslash E$, and given a set of $T$ samples $c^{(1)}, c^{(2)}, ..., c^{(T)}$ obtained via Gibbs sampling from $P(C|e)$, and assuming the Markov chain corresponding to sampling from $C$ is ergodic, then for any $X_i \in X \backslash C, E$ assuming $\hat{P}(X_i|E)$ is defined by Eq.(29), $\hat{P}(X_i|e) \rightarrow P(X_i|e)$ as $T \rightarrow \infty$ .*

**Proof.** By definition:

$$\hat{P}(X_i|e) = \frac{1}{T} \sum_{t=1}^{T} P(X_i|c^{(t)}, e) \qquad (30)$$

Instead of summing over samples, we can rewrite the expression above to sum over all possible tuples $c \in \mathcal{D}(C)$ and group together the samples corresponding to the same tuple instance $c$. Let $q(c)$ denote the number of times a tuple $C = c$ occurs in the set of samples so that $\sum_{c \in \mathcal{D}(C)} q(c) = T$. It is easy to see that:

$$\hat{P}(X_i|e) = \sum_{c \in \mathcal{D}(C)} P(X_i|c, e) \frac{q(c)}{T} \qquad (31)$$

The fraction $\frac{q(c)}{T}$ is a histogram estimator for the posterior marginal $\hat{P}(c|e)$. Thus, we get:

$$\hat{P}(X_i|e) = \sum_{c \in \mathcal{D}(C)} P(X_i|c, e) \hat{P}(c|e) \qquad (32)$$

Since the Markov chain formed by samples from $C$ is ergodic, $\hat{P}(c|e) \rightarrow P(c|e)$ as $T \rightarrow \infty$ and therefore:

$$\hat{P}(X_i|e) \rightarrow \sum_{c \in \mathcal{D}(C)} P(X_i|c, e) P(c|e) = P(X_i|e)$$

$\square$

## 3.3 Complexity

The time and space complexity of generating samples and estimating the posterior marginals via cutset sampling is dominated by the complexity of $JTC$ in line (a) of the algorithm (Figure 3). Only linear amount of additional memory is required to maintain the running sums of $P(C_i|c_{-i}^{(t)}, e)$ and $P(X_i|c^{(t)}, e)$ used in the posterior marginal estimators.

### 3.3.1 SAMPLE GENERATION COMPLEXITY

Clearly, when $JTC$ is applied to the network $\mathcal{B}$ conditioned on all the cutset variables $C$ and evidence variables $E$, its complexity is time and space exponential in the induced width $w$ of the conditioned network. It is $O(N \cdot d^{(w+1)})$ when $C$ is a $w$-cutset (see Definition 2.6).

Using the notion of a $w$-cutset, we can balance sampling and exact inference. At one end of the spectrum we have plain Gibbs sampling where sample generation is fast, requiring linear space, but may have high variance. At the other end, we have exact algorithm requiring time and space exponential in the induced width of the moral graph. In between these two extremes, we can control the time and space complexity using $w$ as follows.

**Theorem 3.3 (Complexity of sample generation)** *Given a network $\mathcal{B}$ over $X$, evidence $E$, and a $w$-cutset $C$, the complexity of generating a new sample is time and space $O(|C| \cdot N \cdot d^{(w+2)})$ where $d$ bounds the variable's domain size and $N = |X|$.*

**Proof.** If $C$ is a $w$-cutset and $d$ is the maximum domain size, then the complexity of computing joint probability $P(c_i, c_{-i}^{(t)}, e)$ over the conditioned network is $O(N \cdot d^{(w+1)})$. Since this operation must be repeated for each $c_i \in \mathcal{D}(C_i)$, the complexity of processing one variable (computing distribution $P(C_i|c_{-i}^{(t)}, e)$) is $O(N \cdot d \cdot d^{(w+1)}) = O(N \cdot d^{(w+2)})$. Finally, since ordered Gibbs sampling requires sampling each variable in the cutset, generating one sample is $O(|C| \cdot N \cdot d^{(w+2)})$. $\quad\square$

### 3.3.2 Complexity of Estimator Computation

The posterior marginals for any cutset variable $C_i \in C$ are easily obtained at the end of sampling process without incurring additional computation overhead. As mentioned earlier, we only need to maintain a running sum of probabilities $P(c_i|c_{-i}^{(t)}, e)$ for each $c_i \in \mathcal{D}(C_i)$. Estimating $P(X_i|e)$, $X_i \in X \backslash C, E$, using Eq.(29) requires computing $P(X_i|c^{(t)}, e)$ once a sample $c^{(t)}$ is generated. In summary:

**Theorem 3.4 (Computing Marginals)** *Given a $w$-cutset $C$, the complexity of computing posteriors for all variables $X_i \in X \backslash E$ using $T$ samples over the cutset variables is $O(T \cdot [|C| + d] \cdot N \cdot d^{(w+1)})$.*

**Proof.** As we showed in Theorem 3.3, the complexity of generating one sample is $O(|C| \cdot N \cdot d^{(w+2)})$. Once a sample $c^{(t)}$ is generated, the computation of the posterior marginals for the remaining variables requires computing $P(X_i|c^{(t)}, e)$ via $JTC(X_i, c^{(t)}, e)$ which is $O(N \cdot d^{(w+1)})$. The combined computation time for one sample is $O(|C| \cdot N \cdot d^{(w+2)} + N \cdot d^{(w+1)}) = O([|C| + d] \cdot N \cdot d^{(w+1)})$. Repeating the computation for T samples, yields $O(T \cdot [|C| + d] \cdot N \cdot d^{(w+1)})$. $\quad\square$

Note that the space complexity of $w$-cutset sampling is bounded by $O(N \cdot d^{(w+1)})$.

### 3.3.3 Complexity of Loop-Cutset

When the cutset $C$ is a loop-cutset, algorithm $JTC$ reduces to belief propagation (Pearl, 1988) that computes the joint distribution $P(C_i, c_{-i}^{(t)}, e)$ in linear time. We will refer to the special case as *loop-cutset sampling* and to the general as *$w$-cutset sampling*.

A loop-cutset is also a $w$-cutset where $w$ equals the maximum number of unobserved parents (upper bounded by the maximum indegree of a node). However, since processing poly-trees is linear even for large $w$, the induced width does not capture its complexity properly. The notion of loop-cutset could be better captured via the hyperwidth of the network (Gottlob, Leone, & Scarello, 1999; Kask et al., 2005). The hyperwidth of a poly-tree is 1 and therefore, a loop-cutset can be defined as a 1-hypercutset. Alternatively, we can express the complexity via the network's input size $M$ which captures the total size of conditional probability tables to be processed as follows:

**Theorem 3.5 (Complexity of loop-cutset sample generation)** *If $C$ is a loop-cutset, the complexity of generating each sample is $O(|C| \cdot d \cdot M)$ where $M$ is the size of the input network.*

**Proof.** When a loop-cutset of a network is instantiated, belief propagation (BP) can compute the joint probability $P(c_i, c_{-i}^{(t)}, e)$ in linear time $O(M)$ (Pearl, 1988) yielding total time and space of $O(|C| \cdot d \cdot M)$ for each sample. □

### 3.4 Optimizing Cutset Sampling Performance

Our analysis of the complexity of generating samples (Theorem 3.3) is overly pessimistic in assuming that the computation of the sampling distribution for each variable in the cutset is independent. While all variables may change a value when moving from one sample to the next, the change occurs one variable at a time in some sequence so that much of the computation can be retained when moving from one variable to the next .

We will now show that sampling all the cutset variables can be done more efficiently reducing the factor of $N \cdot |C|$ in Theorem 3.3 to $(N + |C| \cdot \delta)$ where $\delta$ bounds the number of clusters in the tree decomposition used by $JTC$ that contains any node $C_i \in C$. We assume that we can control the order by which cutset variables are sampled.

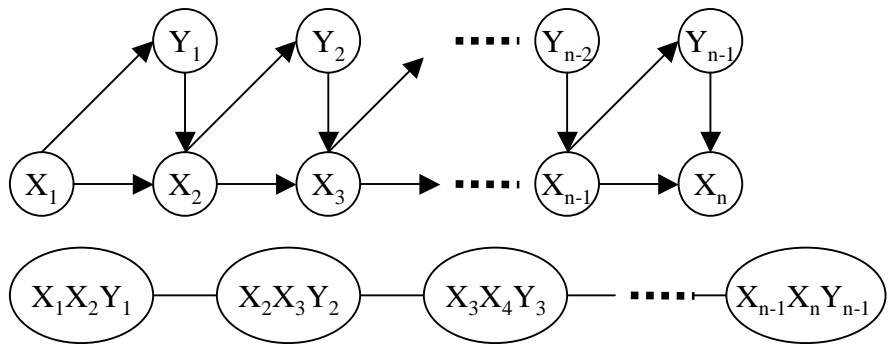

Figure 4: A Bayesian network (top) and corresponding cluster-tree (bottom).

Consider a simple network with variables $X=\{X_1, ....X_n\}$, $Y=\{Y_1, ..., Y_{n-1}\}$ and CPTs $P(X_{i+1}|X_i, Y_i)$ and $P(Y_{i+1}|X_i)$ defined for every $i$ as shown in Figure 4, top. The join-tree of this network is a chain of cliques of size 3 given in Figure 4, bottom. Since $Y$ is a loop-cutset, we will sample variables in $Y$. Let's assume that we use the ordering $Y_1, Y_2, ...Y_{n-1}$ to generate a sample. Given the current sample, we are ready to generate the next sample by applying $JTC$ (or bucket-elimination) to the network whose cutset variables are assigned.

This makes the network effectively singly-connected and leaves only 2 actual variables in each cluster. The algorithm sends a message from the cluster containing $X_n$ towards the cluster containing $X_1$. When cluster $(X_1, X_2, Y_1)$ gets the relevant message from cluster $(X_2, X_3, Y_2)$ we can sample $Y_1$. This can be accomplished by $d$ linear computations in clique $(X_1, X_2, Y_1)$ for each $y_i \in \mathcal{D}(Y_i)$ yielding the desired distribution $P(Y_1|.)$ (we can multiply all functions and incoming messages in this cluster, sum out $X_1$ and $X_2$ and normalize). If the cutset is a $w$-cutset, each computation in a single clique is $O(d^{(w+1)})$.

Once we have $P(Y_1|\cdot)$, $Y_1$ is sampled and assigned a new value, $y_1$. Cluster $(X_1, X_2, Y_1 = y_1)$ then sends a message to cluster $(X_2, X_3, Y_2)$ which now has all the information necessary to compute $P(Y_2|.)$ in $O(d^{(w+2)})$. Once $P(Y_2|.)$ is available, a new value $Y_2 = y_2$ is sampled. The cluster than computes and sends a message to cluster $(X_3, X_4, Y_3)$, and so on. At the end, we obtain a full sample via two message passes over the conditioned network having computation complexity of $O(N \cdot d^{(w+2)})$. This example can be generalized as follows.

**Theorem 3.6** *Given a Bayesian network having $N$ variables, a $w$-cutset $C$, a tree-decomposition $T_r$, and given a sample $c_1, ..., c_{|C|}$, a new sample can be generated in $O((N + |C| \cdot \delta) \cdot d^{(w+2)})$ where $\delta$ is the maximum number of clusters containing any variable $C_i \in C$.*

**Proof.** Given $w$-cutset $C$, by definition, there exists a tree-decomposition $T_r$ of the network (that includes the cutset variables) such that when the cutset variables $C$ are removed, the number of variables remaining in each cluster of $T_r$ is bounded by $w + 1$. Let's impose directionality on $T_r$ starting at an arbitrary cluster that we call $R$ as shown in Figure 5. Let $T_{C_i}$ denote the connected subtree of $T_r$ whose clusters include $C_i$. In Figure 5, for clarity, we collapse the subtree over $C_i$ into a single node. We will assume that cutset nodes are sampled in depth-first traversal order dictated by the cluster tree rooted in $R$.

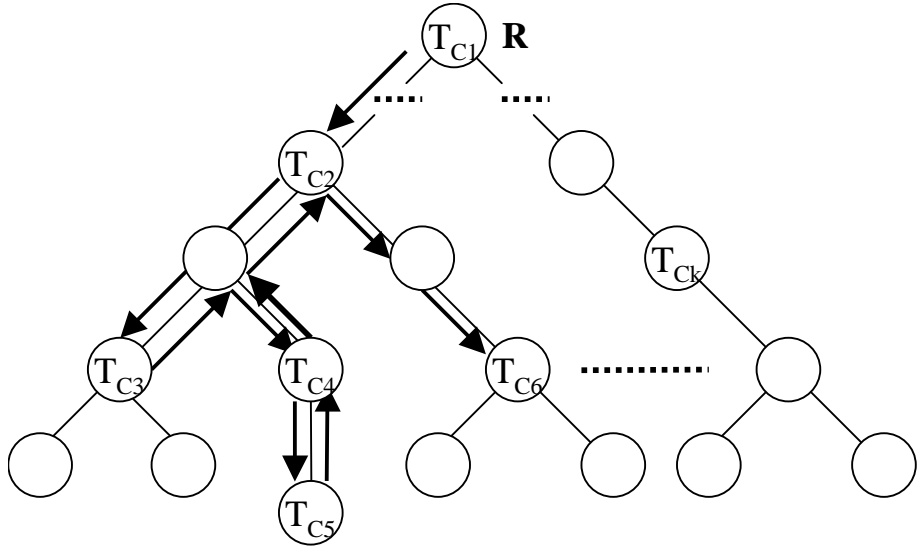

Figure 5: A cluster-tree rooted in cluster $R$ where a subtree over each cutset node $C_i$ is collapsed into a single node marked $T_{C_i}$.

Given a sample $c^{(t)}$, $JTC$ will send messages from leaves of $T_r$ towards the root cluster. We can assume without loss of generality that $R$ contains cutset node $C_1$ which is the first to be sampled in $c^{(t+1)}$. $JTC$ will now pass messages from root down only to clusters restricted to $T_{C_1}$ (note that $R \in T_{C_1}$). Based on these messages $P(C_1 = c_1, c_{-1}^{(t)})$ can be computed in $O(d^{(w+1)})$. We will repeat this computation for each other value of $C_1$ involving only clusters in $T_{C_1}$ and obtain the distribution $P(C_1|\cdot)$ in $O(d^{(w+2)})$ and sample a new value for $C_1$. Thus, if $C_1$ appears in $\delta$ clusters, the number of message passing computations (after the initial $O(N)$ pass) is $O(\delta)$ and we can generate the first distribution $P(C_1|\cdot)$ in $O(\delta \cdot d^{(w+2)})$.

The next node in the depth-first traversal order is $T_{C_2}$ and thus, the second variable to be sampled is $C_2$. The distance between variables $C_1$ and $C_2$, denoted $dist_{1,2}$, is the shortest path along $T_r$ from a cluster that contains $C_1$ to a cluster that contains $C_2$. We apply JTC's mesage-passing along that path only which will take at most $O(dist_{1,2} \cdot d^{(w+1)})$. Then, to obtain the conditional distribution $P(C_2|\cdot)$, we will recompute messages in the subtree of $T_{C_2}$ for each value $c_2 \in \mathcal{D}(C_2)$ in $O(\delta \cdot d^{(w+2)})$. We continue the computation in a similar manner for other cutset nodes.

If $JTC$ traverses the tree in the depth-first order, it only needs to pass messages along each edge twice (see Figure 5). Thus, the sum of all distances traveled is $\sum_{i=2}^{|C|} dist_{i,i-1} = O(N)$. What may be repeated is the computation for each value of the sampled variable. This, however, can be accomplished via message-passing restricted to individual variables' subtrees and is bounded by its $\delta$. We can conclude that a new full sample can be generated in $O((N + |C| \cdot \delta) \cdot d^{(w+2)})$. $\quad\square$

It is worthwhile noting that the complexity of generating a sample can be further reduced by a factor of $d/(d-1)$ (which amounts to a factor of 2 when $d = 2$) by noticing that whenever we move from variable $C_i$ to $C_{i+1}$, the joint probability $P(c_1^{(t+1)}, ..., c_i^{(t+1)}, c_{i+1}^{(t)}, ..., c_k^{(t)})$ is already available from the previous round and should not be recomputed. We only need to compute $P(c_1^{(t+1)}, ..., c_i^{(t+1)}, c_{i+1}, ..., c_k^{(t)})$ for $c_{i+1} \neq c_{i+1}^{(t)}$. Buffering the last computed joint probability, we only need to apply $JTC$ algorithm $d - 1$ times. Therefore, the total complexity of generating a new sample is $O((N + |C| \cdot \delta) \cdot (d-1) \cdot d^{(w+1)})$.

**Example 3.7** *Figure 6 demonstrates the application of the enhancements discussed. It depicts the moral graph (a), already triangulated, and the corresponding join-tree (b) for the Bayesian network in Figure 1. With evidence variable $E$ removed, variables $B$ and $D$ form a 1-cutset. The join-tree of the network with cutset and evidence variables removed is shown in Figure 6 (c). Since removing $D$ and $E$ from cluster $DFE$ leaves only one variable, $F$, we combine clusters $BDF$ and $DFE$ into one cluster, $FG$. Assuming that cutset variables have domains of size 2, we can initialize $B = b^0$ and $D = d^0$.*

*Selecting cluster $AC$ as the root of the tree, $JTC$ first propagates messages from leaves to the root as shown in Figure 6 (c) and then computes $P(b^0, d^0, e)$ in cluster $AC$. Next, we set $B = b^1$; updating all functions containing variable $B$, and propagating messages through the subtree of $B$ consisting of clusters $AC$ and $CF$ (Figure 6 (d)), we obtain $P(b^1, d^0, e)$. Normalizing the two joint probabilities, we obtain distribution $P(B|d^0, e)$ and sample a new value of $B$. Assume we sampled value $b^1$.*

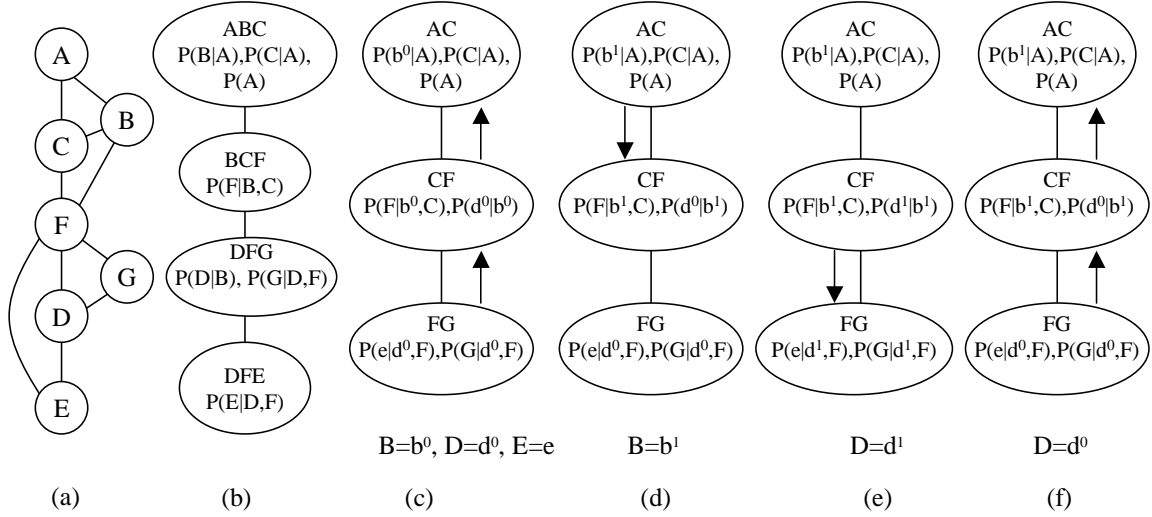

Figure 6: A join-tree of width 2 (b) for a moral graph (a) is transformed into a join-tree of width 1 (c) when evidence variable $E$ and cutset variables $B$ and $D$ are instantiated (in the process, clusters $BDF$ and $BCF$ are merged into cluster $CF$). The clusters contain variables and functions from the original network. The cutset nodes have domains of size 2, $\mathcal{D}(B) = \{b^0, b^1\}$, $\mathcal{D}(D) = \{d^0, d^1\}$. Starting with a sample $\{b^0, d^0\}$, messages are propagated in (c)-(e) to first, sample a new value of variable $B$ (d) and then variable $D$ (e). Then messages are propagated up the tree to compute posterior marginals $P(\cdot|b^1, d^0, e)$ for the rest of the variables (f).

Next, we need to compute $P(D|b^1, e)$ to sample a new value for variable $D$. The joint probability $P(d^0, b^1, e)$ is readily available since it was computed for sampling a new value of $B$. Thus, we set $D = d^1$ and compute the second probability $P(d^1, b^1, e)$ updating functions in clusters $CF$ and $FG$ and sending an updated message from $CF$ to $FG$ (Figure 6 (e)). We obtain distribution $P(D|b^1, e)$ by normalizing the joint probabilities and sample a new value $d^0$ for $D$. Since the value has changed from latest computation, we update again functions in the clusters $CF$ and $FG$ and propagate updated messages in the subtree $C_D$ (send message from $CF$ to $FG$).

In order to obtain the distributions $P(\cdot|b^1, d^0, e)$ for the remaining variables $A$, $C$, $F$, and $G$, we only need to send updated messages up the join-tree, from $FG$ to $CF$ and then from $CF$ to $AC$ as shown in Figure 6 (f). The last step also serves as the initialization step for the next sample generation.

In this example the performance of cutset sampling is significantly better than its worst case. We have sent a total of 5 messages to generate a new sample while the worst case suggests at least $N \cdot |C| \cdot d = 3 \cdot 2 \cdot 2 = 12$ messages (here, $N$ equals the number of clusters).

### 3.5 On finding A $w$-Cutset

Clearly, $w$-cutset sampling will be effective only when the $w$-cutset is small. This calls for the task of finding a minimum size $w$-cutset. The problem is NP-hard; yet, several heuristic algorithms have been proposed. We next briefly survey some of those proposals.

Larossa and Dechter (2003) obtain $w$-cutset when processing variables in the elimination order. The next node to be eliminated (selected using some triangulation heuristics) is added to the cutset if its current induced width (or degree) is greater than $w$. Geiger and Fishelson (2003) agument this idea with various heuristics.

Bidyuk and Dechter (2003) select the variables to be included in the cutset using greedy heuristics based on the node's basic graph properties (such as the degree of a node). One scheme starts from an empty $w$-cutset and then heuristically adds nodes to the cutset until a tree-decomposition of width $\leq w$ can be obtained. The other scheme starts from a set $C = X \backslash E$ containing all nodes in the network as a cutset and then removes nodes from the set in some order. The algorithm stops when removing the next node would result in a tree decomposition of width $> w$.

Alternatively, Bidyuk and Dechter (2004) proposed to first obtain a tree-decomposition of the network and then find the minimal $w$-cutset of the tree-decomposition (also an NP-hard problem) via a well-known greedy algorithm used for set cover problem. This approach is shown to yield a smaller cutset than previously proposed heuristics and is used for finding $w$-cutset in our experiments (section 4.4) with a modification that a tree-decomposition is re-computed each time a node is removed from the tree and added to the $w$-cutset.

## 4. Experiments

In this section, we present empirical studies of cutset sampling algorithms for several classes of problems. We use the mean square error of the posterior marginals' estimates as a measure of accuracy. We compare with traditional Gibbs sampling, likelihood weighting (Fung & Chang, 1989; Shachter & Peot, 1989), and the state of the art AIS-BN adaptive importance sampling algorithm (Cheng & Druzdzel, 2000). We implemented AIS-BN using the parameters specified by Cheng and Druzdzel (2000). By using our own implementation, we made sure that all sampling algorithms used the same data access routines and the same error measures providing a uniform framework for comparing their performance. For reference we also report the performance of Iterative Belief Propagation (IBP) algorithm.

### 4.1 Methodology

In this section we detail describe methodology used and the implementation decisions made that apply to the collection of the empirical results.

#### 4.1.1 SAMPLING METHODOLOGY

In all sampling algorithms we restarted the Markov chain every $T$ samples. The samples from each chain (batch) $m$ are averaged separately:

$$\hat{P}_m(x_i|e) = \frac{1}{T} \sum_{t=1}^{T} P(x_i|c^{(t)}, e)$$

The final estimate is obtained as a sample average over M chains:

$$\hat{P}(x_i|e) = \frac{1}{M}\sum_{m=1}^{M}\hat{P}_m(x_i|e)$$

Restarting a Markov chain is known to improve the sampling convergence rate. A single chain can become "stuck" generating samples from a single high-probability region without ever exploring large number of other high-probability tuples. By restarting a Markov chain at a different random point, a sampling algorithm can achieve a better coverage of the sampling space. In our experiments, we did not observe any significant difference in the estimates obtained from a single chain of size $M \cdot T$ or $M$ chains of size $T$ and therefore, we only choose to report the results for multiple Markov chains. However, we rely on the independence of random values $\hat{P}_m(x_i|e)$ to estimate 90% confidence interval for $\hat{P}(x_i|e)$.

In our implementation of Gibbs sampling schemes, we use zero "burn-in" time (see section 2.3.1). As we mentioned earlier, the idea of burn-in time is to throw away the first $K$ samples to ensure that the remaining samples are drawn from distribution close to target distribution $P(X|e)$. While conservative methods for estimating $K$ through drift and minorization conditions were proposed by Rosenthal (1995) and Roberts and Tweedie (1999, 2001), the required analysis is beyond the scope of this paper. We consider our comparison between Gibbs sampling and cutset sampling, which is the objective, fair in the sense that both schemes use $K=0$. Further, our experimental results showed no positive indication that burn-in time would be beneficial. In practice, burn-in is the "pre-processing" time used by the algorithm to find the high-probability regions in the distribution $P(C|e)$ in case it initially spends disproportionally large period of time in low probability regions. Discarding a large number of low-probability tuples obtained initially, the frequency of the remaining high-probability tuples is automatically adjusted to better reflect their weight.

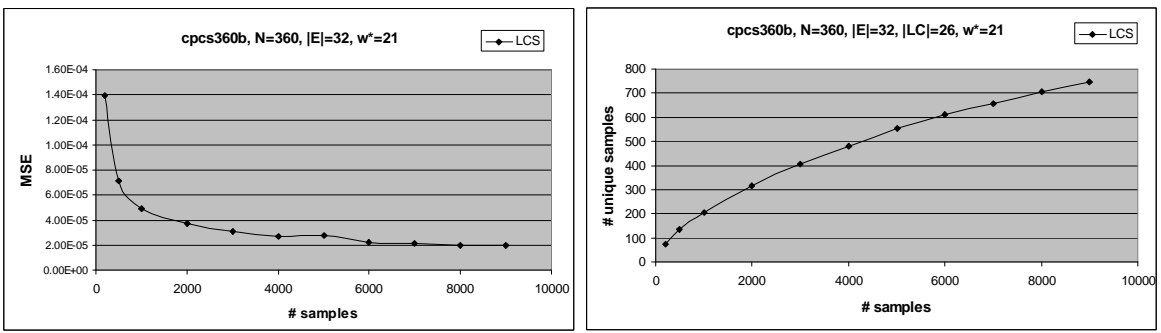

Figure 7: Comparing loop-cutset sampling MSE vs. number of samples (left) and and number of unique samples vs. number of samples (right) in cpcs360b. Results are averaged over 10 instances with different observations.

In our benchmarks, we observed that both full Gibbs sampling and cutset sampling were able to find high probability tuples fast relative to the number of samples generated. For example, in one of the benchmarks, cpcs360b, the rate of generating unique samples, namely, the ratio of cutset instances that have not been seen to the number of samples, decreases over time. Specifically, loop-cutset sampling generates 200 unique tuples after the first 1000 samples, an additional 100 unique tuples while generating the next 1000 samples, and then the rate of generating unique tuples slows to 50 per 1000 samples in the range from 2000 to 10000 samples as shown in Figure 7, right. That means that after the first few hundred samples, the algorithm spends most of the time revisiting high-probability tuples. In other benchmarks, the number of unique tuple instances generated increases linearly (as in cpcs54) and, thus, the tuples appear to be distributed nearly uniformly. In this case, there is no need for burn-in because there are no strongly-expressed heavy-weight tuples. Instead of using burn-in times, we sample initial variable values from the posterior marginal estimates generated by IBP in all of our experiments. Our sampling time includes the pre-processing time of IBP.

All experiments were performed on 1.8 GHz CPU.

### 4.1.2 Measures of Performance

For each problem instance defined by a Bayesian network $\mathcal{B}$ having variables $X = \{X_1, ..., X_n\}$ and evidence $E \subset X$, we derived the exact posterior marginals $P(X_i|e)$ using bucket-tree elimination (Dechter, 2003, 1999a) and computed the mean square error (MSE) of the approximate posterior marginals $\hat{P}(X_i|e)$ for each algorithm where MSE is defined by:

$$MSE = \frac{1}{\sum_{X_i \in X \setminus E} |\mathcal{D}(X_i)|} \sum_{X_i \in X \setminus E} \sum_{\mathcal{D}(X_i)} [P(x_i|e) - \hat{P}(x_i|e)]^2$$

While the mean square error is our primary accuracy measure, the results are consistent across other well-known measures such as average absolute error, KL-distance, and squared Hellinger's distance which we show only for loop-cutset sampling. The absolute error $\Delta$ is averaged over all values of all unobserved variables:

$$\Delta = \frac{1}{\sum_{X_i \in X \setminus E} |\mathcal{D}(X_i)|} \sum_{X_i \in X \setminus E} \sum_{\mathcal{D}(X_i)} |P(x_i|e) - \hat{P}(x_i|e)|$$

KL-distance $D_K$ between the distribution $P(X_i|e)$ and the estimator $\hat{P}(X_i|e)$ is defined as follows:

$$D_K(P(X_i|e), \hat{P}(X_i|e)) = \sum_{\mathcal{D}(X_i)} P(x_i|e) \lg \frac{P(x_i|e)}{\hat{P}(x_i|e)}$$

For each benchmark instance, we compute the KL-distance for each variable $X_i \in X \setminus E$ and then average the results:

$$D_K(P, \hat{P}) = \frac{1}{|X \setminus E|} \sum_{X_i \in X \setminus E} D_K(P(X_i|e), \hat{P}(X_i|e))$$

The squared Hellinger's distance $D_H$ between the distribution $P(X_i|e)$ and the estimator $\hat{P}(X_i|e)$ is obtained as:

$$D_H(P(X_i|e), \hat{P}(X_i|e)) = \sum_{\mathcal{D}(X_i)} [\sqrt{P(x_i|e)} - \sqrt{\hat{P}(x_i|e)}]^2$$

The average squared Hellinger's distance for a benchmark instance is the average of the distances between posterior distributions of one variable:

$$D_H(P, \hat{P}) = \frac{1}{|X \backslash E|} \sum_{X_i \in X \backslash E} D_H(P(X_i|e), \hat{P}(X_i|e))$$

The average errors for different network instances are then averaged over all instances of the given network (typically, 20 instances).

We also report the confidence interval for the estimate $\hat{P}(x_i|e)$ using an approach similar to the well-known batch means method (Billingsley, 1968; Geyer, 1992; Steiger & Wilson, 2001). Since chains are restarted independently, the estimates $\hat{P}_m(x_i|e)$ are independent. Thus, the confidence interval can be obtained by measuring the variance in the estimators $\hat{P}(X_i|e)$. We report results in Section 4.5.

## 4.2 Benchmarks

We experimented with four classes of networks:

**CPCS**. We considered four CPCS networks derived from the Computer-based Patient Case Simulation system (Parker & Miller, 1987; Pradhan, Provan, Middleton, & Henrion, 1994). CPCS network representation is based on INTERNIST 1 (Miller, Pople, & Myers, 1982) and Quick Medical Reference (QMR) (Miller, Masarie, & Myers, 1986) expert systems. The nodes in CPCS networks correspond to diseases and findings and conditional probabilities describe their correlations. The **cpcs54** network consists of $N$=54 nodes and has a relatively large loop-cutset of size $|LC|$=16 ($> 25\%$ of the nodes). Its induced width is 15. **cpcs179** network consists of $N$=179 nodes. Its induced width is $w^*$=8. It has a small loop-cutset of size $|LC|$=8 but with a relatively large corresponding adjusted induced width $w_{LC}$=7. The **cpcs360b** is a larger CPCS network with 360 nodes, adjusted induced width of 21, and loop-cutset $|LC|$=26. Exact inference on cpcs360b averaged $\sim 30$ minutes. The largest network, **cpcs422b**, consisted of 422 nodes with induced width $w^*$=22 and loop-cutset of size 47. The exact inference time for cpcs422b is about 50 minutes.

**Hailfinder network**. It is a small network with only 56 nodes. The exact inference in Hailfinder network is easy since its loop-cutset size is only 5. Yet, this network has some zero probabilities and, therefore, is a good benchmark for demonstrating the convergence of cutset sampling in contrast to Gibbs sampling.

**Random networks**. We experimented with several classes of random networks: random networks, 2-layer networks, and grid networks. The **random networks** were generated with $N$=200 binary nodes (domains of size 2). The first 100 nodes, $\{X_1, ..., X_{100}\}$, were designated as root nodes. Each non-root node $X_i$, $i > 100$, was assigned 3 parents selected randomly from the list of predecessors $\{X_1, ..., X_{i-1}\}$. We will refer to this class of random networks as *multi-partite* random networks to distinguish from bi-partite (2-layer) random networks. The **random 2-layer** networks were generated with 50 root nodes (first layer) and 150 leaf nodes (second layer), yielding a total of 200 nodes. A sample 2-layer random network is shown in Figure 8, left. Each non-root node (second layer) was assigned 1-3 parents selected at random among the root nodes. All nodes were assigned a domain of size 2, $\mathcal{D}(X_i) = \{x_i^0, x_i^1\}$.

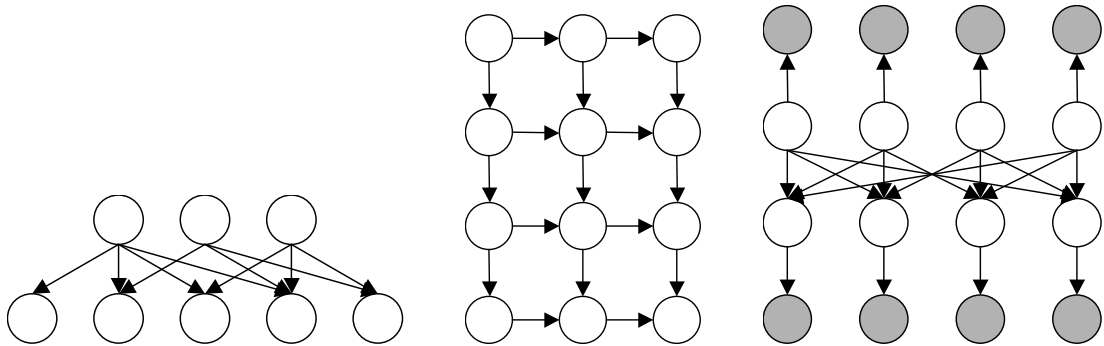

Figure 8: Sample random networks: 2-layer (left), grid (center), coding (right).

For both 2-layer and multi-partite random networks, the root nodes were assigned uniform priors while conditional probabilities were chosen randomly. Namely, each value $P(x_i^0|pa_i)$ was drawn from uniform distribution over interval $(0, 1)$ and used to compute the complementary probability value $P(x_i^1|pa_i) = 1 - P(x_i^0|pa_i)$.

The directed **grid networks** (as opposed to grid-shaped undirected Markov Random Fields) of size 15x30 with 450 nodes were also constructed with uniform priors (on the single root node) and random conditional probability tables (as described above). A sample grid network is shown in Figure 8, center. Those networks had an average induced width of size 20 (exact inference using bucket elimination required about 30 minutes). They had the most regular structure of all and the largest loop-cutset containing nearly a half of all the unobserved nodes.

**Coding networks**. We experimented with coding networks with 50 code bits and 50 parity check bits. The parity check matrix was randomized; each parity check bit had three parents. A sample coding network with 4 code bits, 4 parity checking bits, and total of 8 transmitted bits is shown in Figure 8, center. The total number of variables in each network in our experiments was 200 (50 code bits, 50 parity check bits, and 1 transmitted bit for each code or parity check bit). An average loop-cutset size was 26 and induced width was 21. The Markov chain produced by Gibbs sampling over the whole coding network is not ergodic due to the deterministic parity check function. As a result, Gibbs sampling does not converge. However, the Markov chain corresponding to sampling the subspace of coding bits only is ergodic and, thus, all of the cutset sampling schemes have converged as we will show in the next two sections.

In all networks, except coding and grid networks, evidence nodes were selected at random among the leaf nodes (nodes without children). Since a grid network has only one leaf node, the evidence in the grid networks was selected at random among all nodes. For each benchmark, we report on the chart title the number of nodes in the network $N$, average number of evidence nodes $|E|$, size of loop-cutset $|LC|$, and average induced width of the input instance denoted $w^*$ to distinguish from the induced width $w$ of the network adjusted for its $w$-cutset.

### 4.3 Results for Loop-Cutset Sampling

In this section we compare loop-cutset sampling with pure Gibbs sampling, likelihood weighting, AIS-BN, and IBP. In all benchmarks, the cutset was selected so that the evidence and sampling nodes together constitute a loop-cutset of the network using the algorithm proposed by Becker et al. (2000). We show the accuracy of Gibbs and loop-cutset sampling as a function of the number of samples and time.

**CPCS networks.** The results are summarized in Figures 9-12. The loop-cutset curve in each chart is denoted LCS (for Loop Cutset Sampling). The induced width of the network $w_{LC}$ when loop-cutset nodes are observed is specified in the caption. It is identical to the largest family size in the poly-tree generated when cutset variables are removed. We plot the time on the $x$-axis and the accuracy (MSE) on the $y$-axis. In the CPCS networks, IBP always converged and converged fast (within seconds). Consequently, IBP curve is always a straight horizontal line as the results do not change after the convergence is achieved. The curves corresponding to Gibbs sampling, loop-cutset sampling, likelihood weighting, and AIS-BN demonstrate the convergence of the sampling schemes with time. In the three CPCS networks loop-cutset sampling converges much faster than Gibbs sampling. The only exception is cpcs422b (Figure 12, right) where the induced width of the conditioned singly-connected network remains high ($w_{LC} = 14$) due to large family sizes and thus, loop-cutset sampling generates samples very slowly (4 samples/second) compared to Gibbs sampling (300 samples/second). Since computing sampling distribution is exponential in $w$, sampling a single variable is $O(2^{14})$ (all variables have domains of size 2). As a result, although loop-cutset sampling shows a significant reduction in MSE as a function of the number of samples (Figure 12, left), it is not enough to compensate for the two orders of magnitude difference in the loop-cutset rate of sample generation. For cpcs54 (Figure 9), cpcs179 (Figure 10), and cpcs360b (Figure 11) loop-cutset sampling achieves greater accuracy than IBP within 10 seconds or less.

In comparison with importance sampling schemes, we observe that the AIS-BN algorithm consistently outperforms likelihood weighting and AIS-BN is slightly better than loop-cutset sampling in cpcs54, where the probability of evidence $P(e)=0.0928$ is relatively high. In cpcs179, where probability of evidence $P(e)=4E-05$ is smaller, LCS outperforms AIS-BN while Gibbs sampling curves falls in between AIS-BN and likelihood weighting. Both Gibbs sampling and loop-cutset sampling outperform AIS-BN in cpcs360b and cpcs422b where probability of evidence is small. In cpcs360b average $P(e)=5E-8$ and in cpcs422b the probability of evidence varies from 4E-17 to 8E-47. Note that likelihood weighting and AIS-BN performed considerably worse than either Gibbs sampling or loop-cutset sampling in all of those benchmarks as a function of the number of samples. Consequently, we left them off the charts showing the convergence of Gibbs and loop-cutset sampling as a function of the number of samples in order to zoom in on the two algorithms which are the focus of the empirical studies.

**Coding Networks.** The results for coding networks are shown in Figure 13. We computed error measures over all coding bits and averaged over 100 instances (10 instances, with different observed values, of each of the 10 networks with different coding matrices). As we noted earlier, the Markov chains corresponding to Gibbs sampling over coding networks are not ergodic and, as a result, Gibbs sampling does not converge. However, the Markov chain corresponding to sampling the subspace of code bits only is ergodic and therefore, loop-cutset sampling, which samples a subset of coding bits, converges and even achieves higher accuracy than IBP with time. In reality, IBP is certainly preferable for coding networks since the size of the loop-cutset grows linearly with the number of code bits.

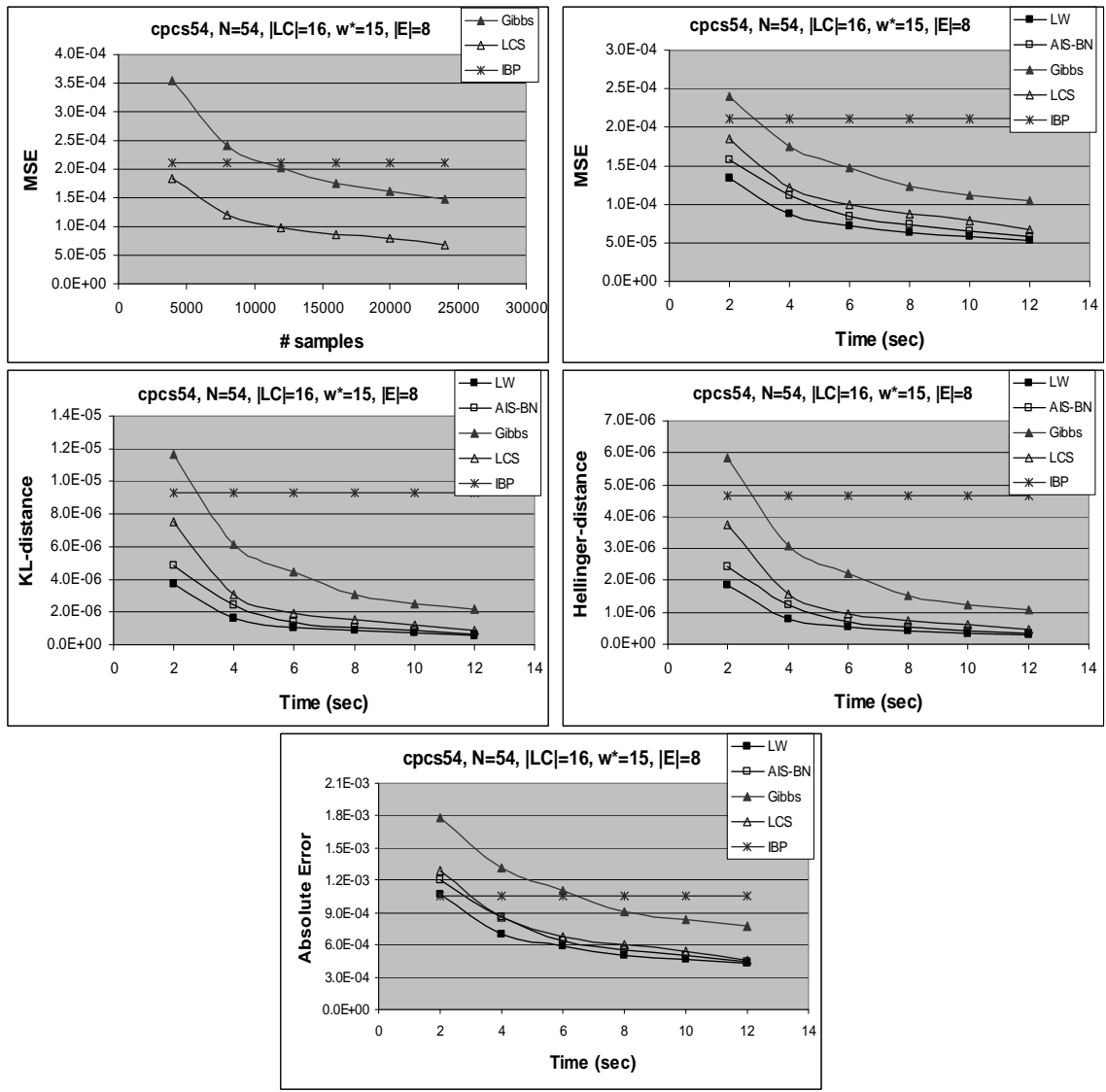

Figure 9: Comparing loop-cutset sampling (LCS), $w_{LC}$=5, Gibbs sampling (hereby referred to as Gibbs), likelihood weighting (LW), AIS-BN, and IBP on cpcs54 network, averaged over 20 instances, showing MSE as a function of the number of samples (top left) and time (top right) and KL-distance (middle left), squared Hellinger's distance (middle right), and an average error (bottom) as a function of time.

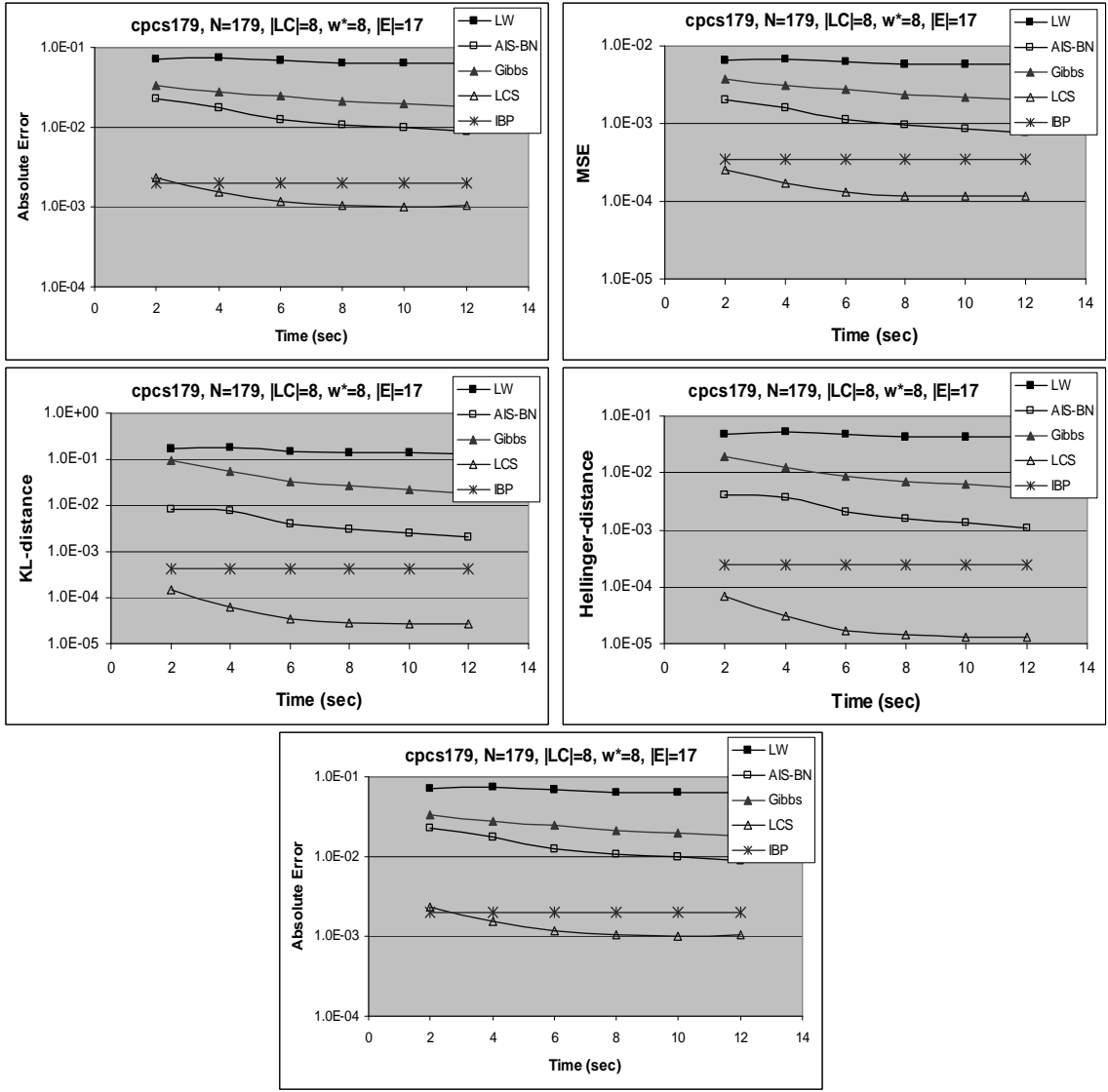

Figure 10: Comparing loop-cutset sampling (LCS), $w_{LC}=7$, Gibbs sampling, likelihood weighting (LW), AIS-BN, and IBP on cpcs179 network, averaged over 20 instances, showing MSE as a function of the number of samples (top left) and time (top right) and KL-distance (middle left), squared Hellinger's distance (middle right), and an average error (bottom) as a function of time.

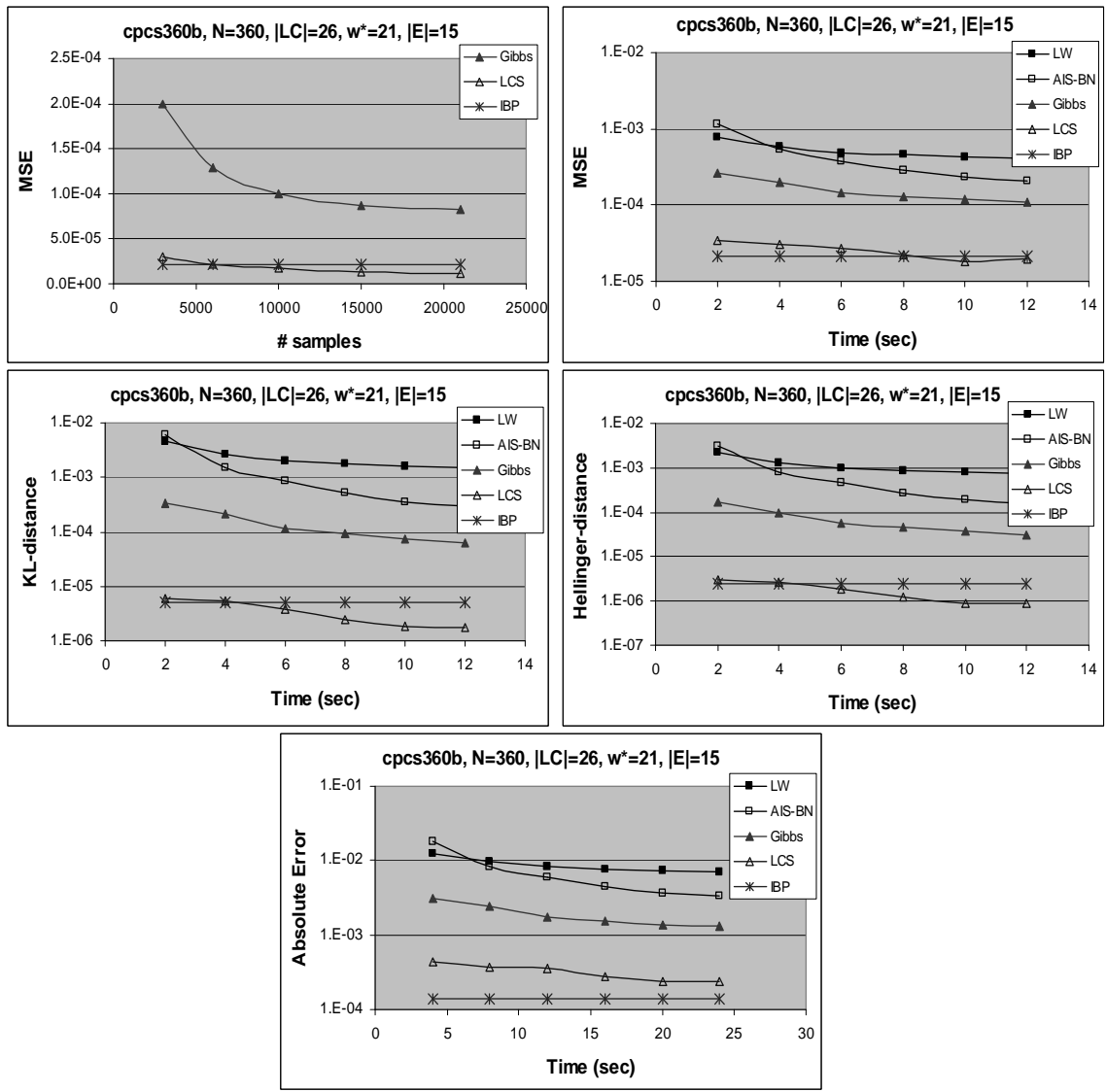

Figure 11: Comparing loop-cutset sampling (LCS), $w_{LC}$=3, Gibbs sampling, likelihood weighting (LW), AIS-BN, and IBP on cpcs360b network, averaged over 20 instances, showing MSE as a function of the number of samples (top left) and time (top right) and KL-distance (middle left), squared Hellinger's distance (middle right), and an average error (bottom) as a function of time.

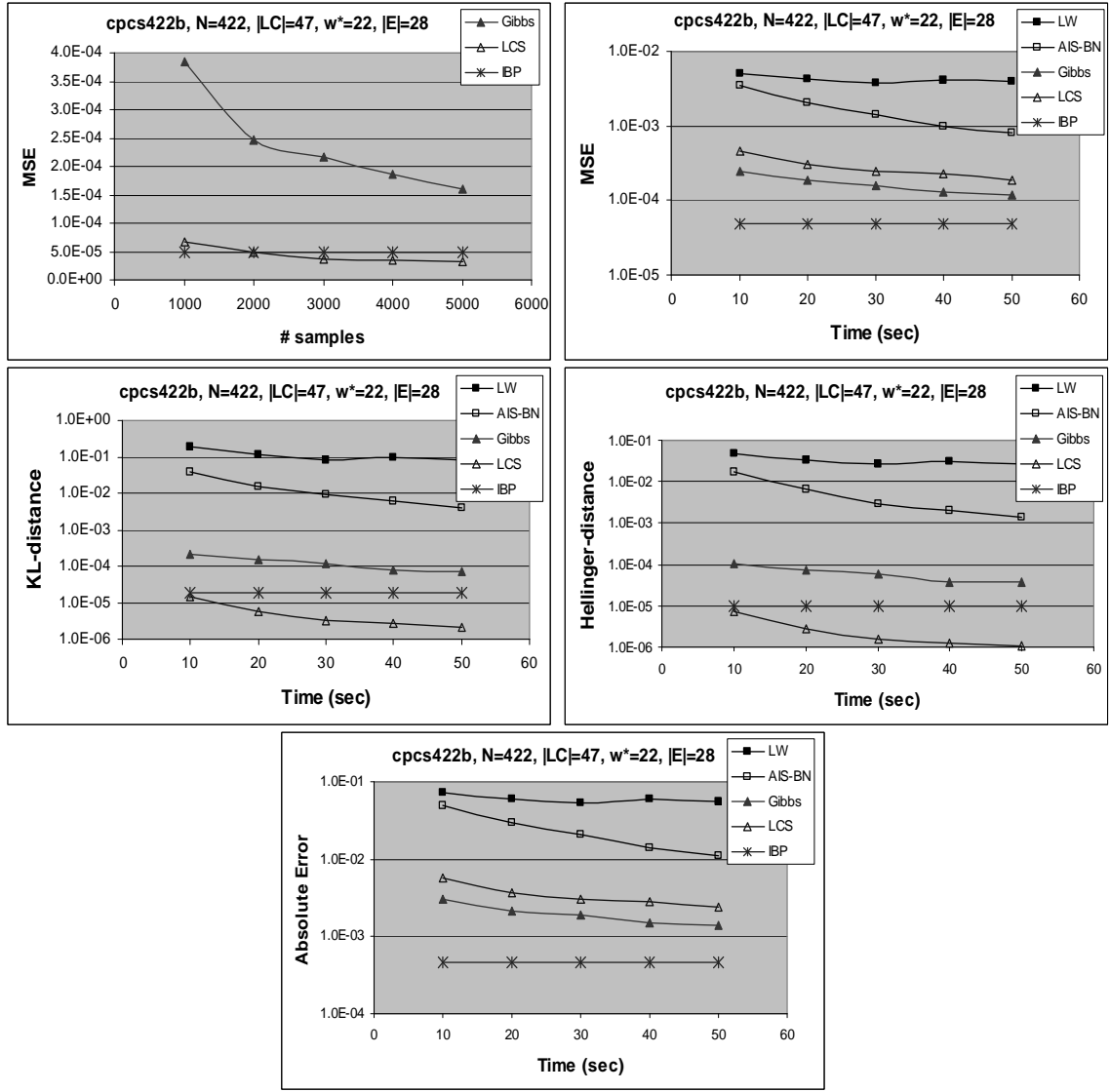

Figure 12: Comparing loop-cutset sampling (LCS), $w_{LC}$=14, Gibbs sampling, likelihood weighting (LW), AIS-BN sampling, and IBP on cpcs422b network, averaged over 10 instances, showing MSE as a function of the number of samples (top left) and time (top right) and KL-distance (middle left), squared Hellinger's distance (middle right), and an average error (bottom) as a function of time.

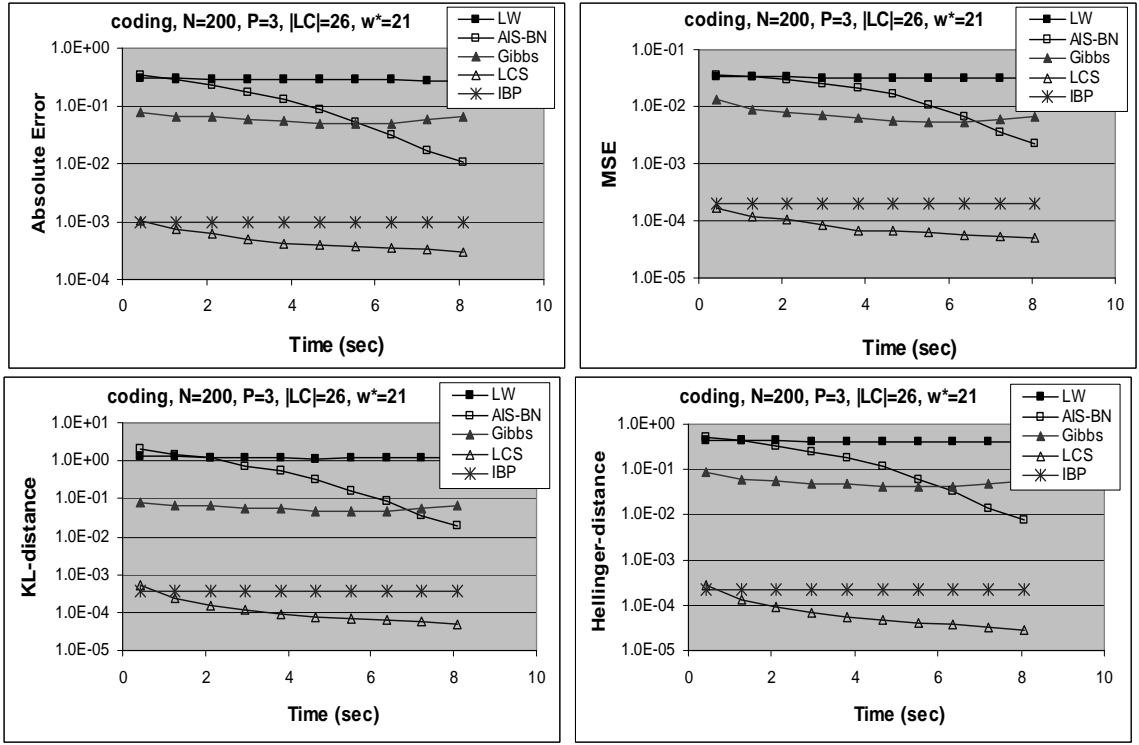

Figure 13: Comparing loop-cutset sampling (LCS), $w_{LC}$=3, Gibbs sampling, likelihood weighting (LW), AIS-BN, and IBP on coding networks, $\sigma$=0.4, averaged over 10 instances of 10 coding networks (100 instances total). The graphs show average absolute error ( top left), MSE (top right), KL-distance (bottom left), and squared Hellinger's distance (bottom right) as a function of time.

**Random networks.** In random multi-part networks (Figure 14, top) and random 2-layer networks (Figure 14, middle), loop-cutset sampling always converged faster than Gibbs sampling. The results are averaged over 10 instances of each network type. In both cases, loop-cutset sampling achieved accuracy of IBP in 2 seconds or less. In 2-layer networks, Iterative Belief Propagation performed particularly poorly. Both Gibbs sampling and loop-cutset sampling obtained more accurate results in less than a second.

**Hailfinder network**. We used this network (in addition to coding networks) to compare the behavior of cutset sampling and Gibbs sampling in deterministic networks. Since Hailfinder network contains many deterministic probabilities, the Markov chain corresponding to Gibbs sampling over all variables is non-ergodic. As expected, Gibbs sampling fails while loop-cutset sampling computes more accurate marginals (Figure 15).

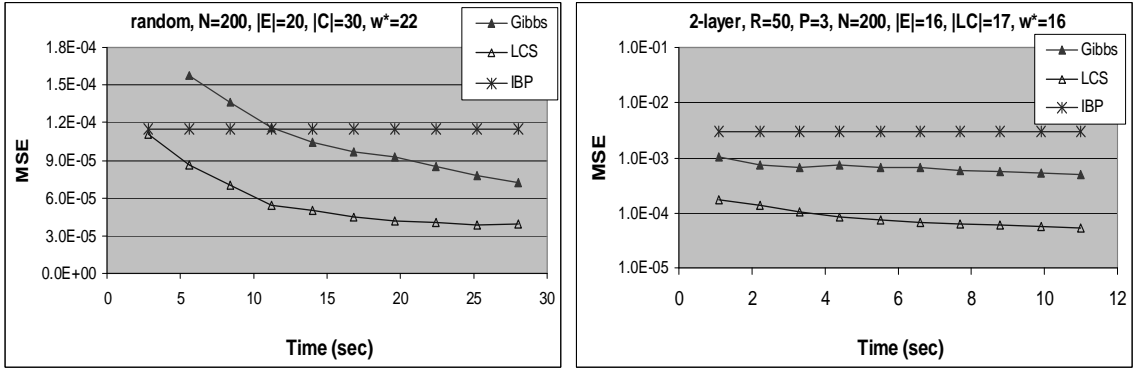

Figure 14: Comparing loop-cutset sampling (LCS), Gibbs sampling, and IBP on random networks (left) and 2-layer random networks (right), $w_{LC}$=3 in both classes of networks, averaged over 10 instances each. MSE as a function of time.

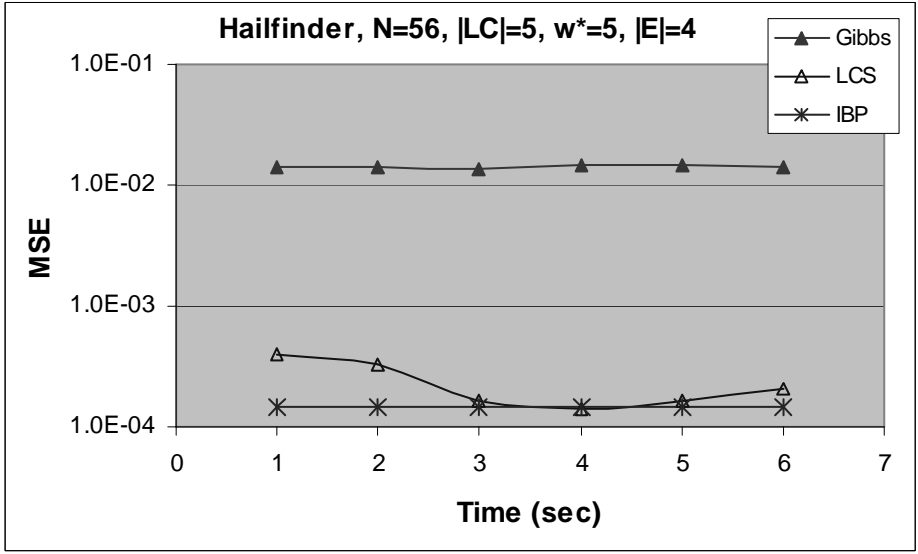

Figure 15: Comparing loop-cutset sampling (LCS), $w_{LC}$=7, Gibbs sampling, and IBP on Hailfinder network, 10 instances. MSE as a function of time.

In summary, the empirical results demonstrate that loop-cutset sampling is cost-effective time-wise and superior to Gibbs sampling. We measured the ratio $R = \frac{M_g}{M_c}$ of the number of samples $M_g$ generated by Gibbs to the number of samples $M_c$ generated by loop-cutset sampling in the same time period (it is relatively constant for any given network and only changes slightly between problem instances that differ with observations). For cpcs54, cpcs179, cpcs360b, and cpcs422b the ratios were correspondingly 2.5, 3.75, 0.7, and 75 (see Table 2 in section 4.4). We also obtained $R$=2.0 for random networks and R=0.3 for random 2-layer networks. The ratio values > 1 indicate that the Gibbs sampler generates samples faster than loop-cutset sampling which is usually the case. In those instances, variance reduction compensated for the increased computation time because fewer samples are needed to converge resulting in the overall better performance of loop-cutset sampling compared to Gibbs sampling. In some cases, however, the reduction in the sample size also compensates for the overhead computation in the sampling of one variable value. In such cases, loop-cutset sampling generated samples faster than Gibbs yielding ratio $R < 1$. Then, the improvement in the accuracy is due both to larger number of samples and to faster convergence.

### 4.4 $w$-Cutset Sampling

In this section, we compare the general $w$-cutset scheme for different values of $w$ against Gibbs sampling. The main goal is to study how the performance of $w$-cutset sampling varies with $w$. For completeness sake, we include results of loop-cutset sampling shown in section 4.3.

In this empirical study, we used the greedy algorithm for set cover problem, mentioned in section 3.5, for finding minimal $w$-cutset. We apply the algorithm in such a manner that each $(w + 1)$-cutset is a proper subset of a $w$-cutset and, thus, can be expected to have a lower variance and converge faster than sampling on $w$-cutset in terms of number of samples required (following the Rao-Blackwellisation theory). We focus the empirical study on the trade-offs between cutset size reduction and the associated increase in sample generation time as we gradually increase the bound $w$.

We used the same benchmarks as before and included also grid networks. All sampling algorithms were given a fixed time bound. When sampling small networks, such as cpcs54 ($w$*=15) and cpcs179 ($w$*=8), where exact inference is easy, sampling algorithms were allocated 10 seconds and 20 seconds respectively. For larger networks we allocated 100-200 seconds depending on the complexity of the network which was only a fraction of exact computation time.

Table 1 reports the size of the sampling set used by each algorithm where each column reports the size of the corresponding $w$-cutset. For example, for cpcs360b, the average size of Gibbs sample (all nodes except evidence) is 345, the loop-cutset size is 26, the size of 2-cutset is 22, and so on. Table 2 shows the rate of sample generation by different algorithms per second. As we observed previously in the case of loop-cutset sampling, in some special cases cutset sampling generated samples faster than Gibbs sampler. For example, for cpcs360b, loop-cutset sampling and 2-cutset sampling generated 600 samples per second while the Gibbs sampler was able to generate only 400 samples. We attribute this to the size of cutset sample (26 nodes or less as reported in Table 1) compared to the size of the Gibbs sample (over 300 nodes).

**CPCS networks.** We present two charts. One chart demonstrates the convergence over time for several values of $w$. The second chart depicts the change in the quality of approximation (MSE) as a function of $w$ for two time points, at the half of the total sampling time and at the end of total sampling time. The performance of Gibbs sampling and cutset sampling for cpcs54 is shown in Figure 16. The results are averaged over 20 instances with 5-10 evidence variables. The graph on the left in Figure 16 shows the mean square error of the estimated posterior marginals as a function of time for Gibbs sampling, loop-cutset sampling, and $w$-cutset sampling for $w$=2, 3, 4, and 5. The second chart shows accuracy as a function of $w$. The first point corresponds to Gibbs sampling; other points correspond to loop-cutset sampling and $w$-cutset sampling with $w$ ranging from 2 to 6. The loop-cutset result is embedded with the $w$-cutset values at $w$=5. As explained in section 3.3, the loop-cutset corresponds to the $w$-cutset where $w$ is the maximum number of parents in the network. Initially, the best results were obtained by 2- and 3-cutset sampling followed by the loop-cutset sampling. With time, 2- and 5-cutset sampling become the best.

| | **Sampling Set Size** | | | | | | | | |
|---|---|---|---|---|---|---|---|---|---|
| | Gibbs | LC | $w$=2 | $w$=3 | $w$=4 | $w$=5 | $w$=6 | $w$=7 | $w$=8 |
| cpcs54 | 51 | 16 | 17 | 15 | 11 | 9 | 8 | - | - |
| cpcs179 | 162 | 8 | 11 | 9 | 7 | 5 | - | - | - |
| cpcs360b | 345 | 26 | 22 | 19 | 16 | 15 | 14 | 13 | - |
| cpcs422b | 392 | 47 | 65 | 57 | 50 | 45 | 40 | 35 | - |
| grid15x30 | 410 | 169 | 163 | 119 | 95 | 75 | 60 | 50 | 13 |
| random | 190 | 30 | 61 | 26 | 25 | 24 | 18 | 17 | - |
| 2layer | 185 | 17 | 22 | 15 | 13 | 13 | 11 | - | - |
| coding | 100 | 26 | 38 | 23 | 18 | 18 | - | - | - |

Table 1: Markov chain sampling set size as a function of $w$.

| | **No. of Samples** | | | | | | | | |
|---|---|---|---|---|---|---|---|---|---|
| | Gibbs | LC | $w$=2 | $w$=3 | $w$=4 | $w$=5 | $w$=6 | $w$=7 | $w$=8 |
| cpcs54 | 5000 | 2000, $w$= 5 | 3000 | 2400 | 800 | 500 | 300 | - | - |
| cpcs179 | 1500 | 400, $w$= 7 | 400 | 150 | 40 | 10 | - | - | - |
| cpcs360b | 400 | 600, $w$= 3 | 600 | 400 | 160 | 100 | 40 | 20 | - |
| cpcs422b | 300 | 4, $w$=14 | 200 | 150 | 90 | 50 | 30 | 15 | - |
| grid15x30 | 2000 | 500, $w$= 2 | 300 | 260 | 150 | 105 | 60 | 35 | 20 |
| random | 2000 | 1000, $w$= 3 | 1400 | 700 | 450 | 300 | 140 | 75 | - |
| 2layer | 200 | 700, $w$= 3 | 900 | 320 | 150 | 75 | 40 | - | - |
| coding | 2400 | 1000, $w$= 3 | 1000 | 400 | 200 | 120 | 100 | - | - |

Table 2: Average number of samples generated per second as a function of $w$.

The results for cpcs179 are reported in Figure 17. Both charts show that loop-cutset sampling and $w$-cutset sampling for $w$ in range from 2 to 5 are superior to Gibbs sampling. The chart on the left shows that the best of the cutset sampling schemes, having the lowest MSE curves, are 2- and 3-cutset sampling. The loop-cutset curve falls in between 2- and 3-cutset at first and is outperformed by both 2- and 3-cutset after 12 seconds. Loop-cutset sampling and 2- and 3-cutset sampling outperform Gibbs sampling by nearly two orders of magnitude as their MSE falls below 1E-04 while Gibbs MSE remains on the order of 1E-02. The 4- and 5-cutset sampling results fall in between, achieving the MSE $\approx$1E-03. The curves corresponding to loop-cutset sampling and 2-, 3- and 4-cutset sampling fall below the IBP line which means that all four algorithms outperform IBP in the first seconds of execution (IBP converges in less than a second). The 5-cutset outperforms IBP after 8 seconds. In Figure 17 on the right, we see the accuracy results for all sampling algorithms

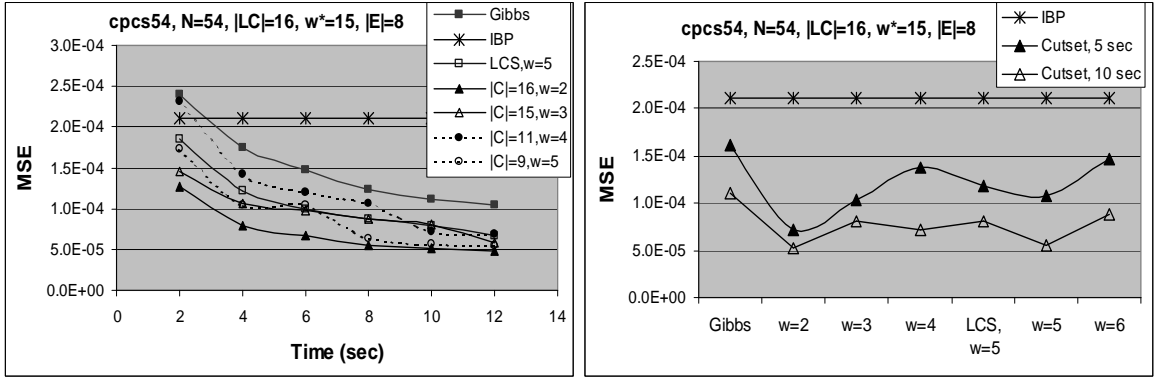

Figure 16: MSE as a function of time (left) and $w$ (right) in cpcs54, 20 instances, time bound=12 seconds.

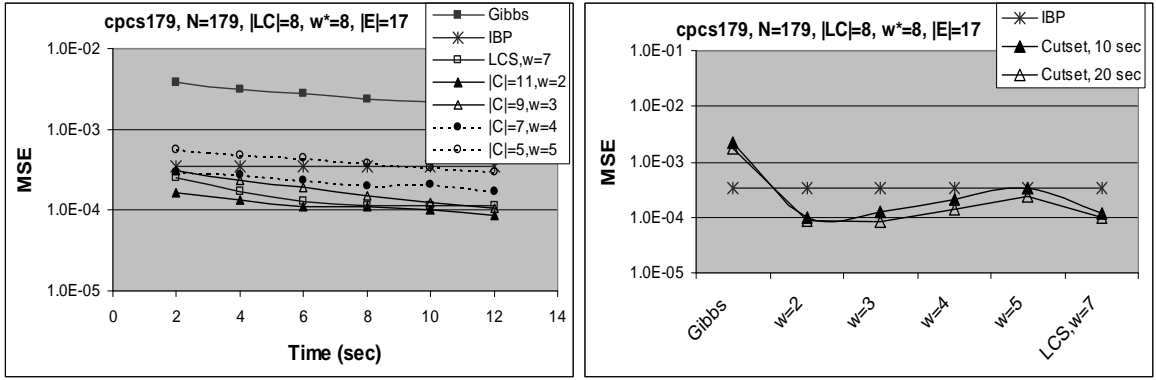

Figure 17: MSE as a function of time (left) and $w$ (right) in cpcs179, 20 instances, time bound=12 seconds. Y-scale is exponential due to large variation in performance of Gibbs and cutset sampling.

after 10 seconds and 20 seconds. They are in agreement with the convergence curves on the left.

In cpcs360b (Figure 18), loop-cutset sampling and 2- and 3-cutset sampling have similar performance. The accuracy of the estimates slowly degrades as $w$ increases. Loop-cutset sampling and $w$-cutset sampling substantially outperform Gibbs sampling for all values $w$ and exceed the accuracy of IBP within 1 minute.

On the example of cpcs422b, we demonstrate the significance of the adjusted induced width of the conditioned network in the performance of cutset sampling. As we reported in section 4.3, its loop-cutset is relatively small $|LC|=47$ but $w_{LC}=14$ and thus, sampling just one new loop-cutset variable value is exponential in the big adjusted induced width. As a result, loop-cutset sampling computes only 4 samples per second while the 2-, 3- and 4-cutset, which are only slightly larger having 65, 57, and 50 nodes respectively (see Table 1), compute samples at rates of 200, 150, and 90 samples per second (see Table 2).

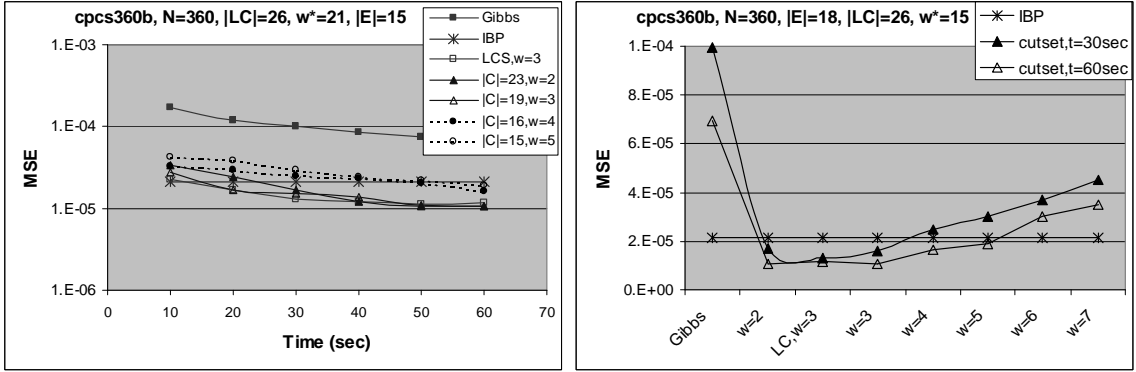

Figure 18: MSE as a function of time (left) and $w$ (right) in cpcs360b, 20 instances, time bound=60 seconds. Y-scale is exponential due to large variation in performance of Gibbs and cutset sampling.

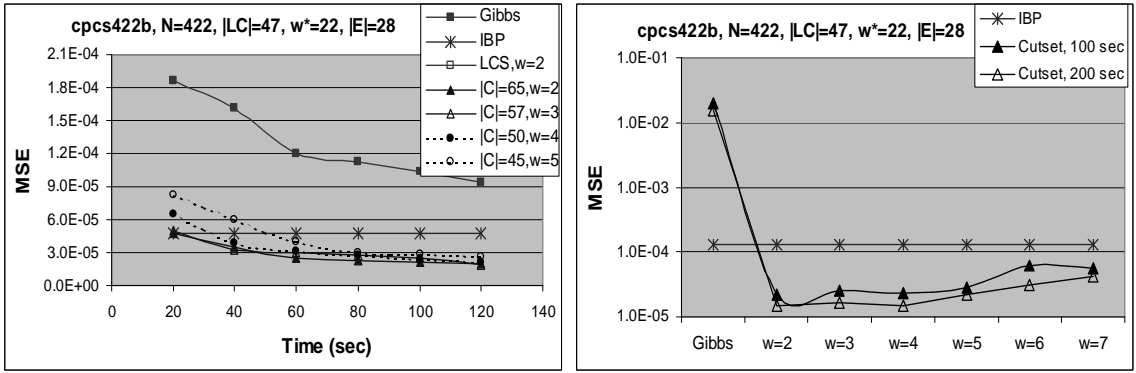

Figure 19: MSE as a function of time (left) and $w$ (right) in cpcs422b, 10 instances, time bound=200 seconds. Y-scale is exponential due to large variation in performance of Gibbs and cutset sampling.

The 5-cutset that is closest to loop-cutset in size, $|C_5| = 45$, computes 50 samples per second which is an order of magnitude more than loop-cutset sampling. The results for cpcs422b are shown in Figure 19. The loop-cutset sampling results are excluded due to its poor performance. The chart on the right in Figure 19 shows that $w$-cutset performed well in range of $w = 2 - 7$ and is far superior to Gibbs sampling. When allowed enough time, $w$-cutset sampling outperformed IBP as well. The IBP converged in 5 seconds. The 2-, 3-, and 4-cutset improved over IBP within 30 seconds, and 5-cutset after 50 seconds.

**Random networks.** Results from 10 instances of random multi-partite and 10 instances of 2-layer networks are shown in Figure 20. As we can see, $w$-cutset sampling substantially improves over Gibbs sampling and IBP reaching optimal performance for $w = 2 - 3$ for both classes of networks. In this range, its performance is similar to that of loop-cutset sampling. In case of 2-layer networks, the accuracy of both Gibbs sampling and

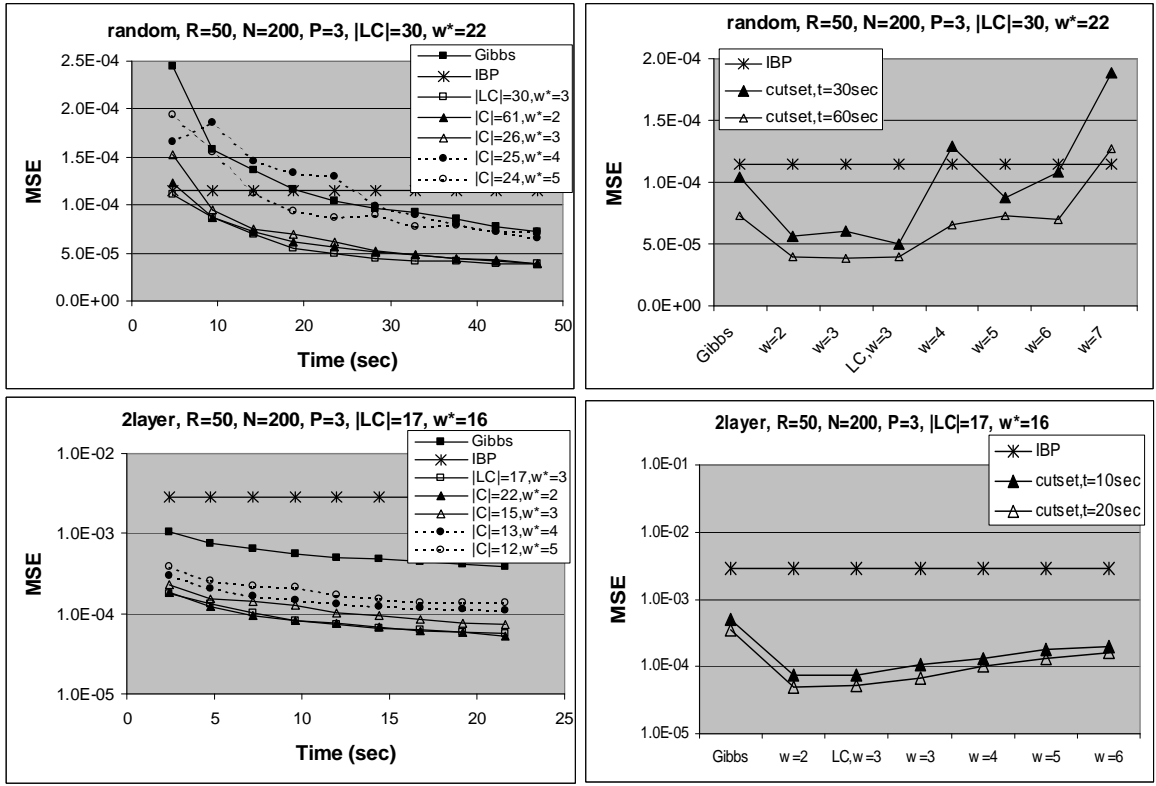

Figure 20: Random multi-partite networks (top) and 2-layer networks (bottom), 200 nodes, 10 instances. MSE as a function of the number of samples (left) and $w$ (right).

IBP is an order-of-magnitude less compared to cutset sampling (Figure 20, bottom right). The poor convergence and accuracy of IBP on 2-layer networks was observed previously (Murphy et al., 1999).

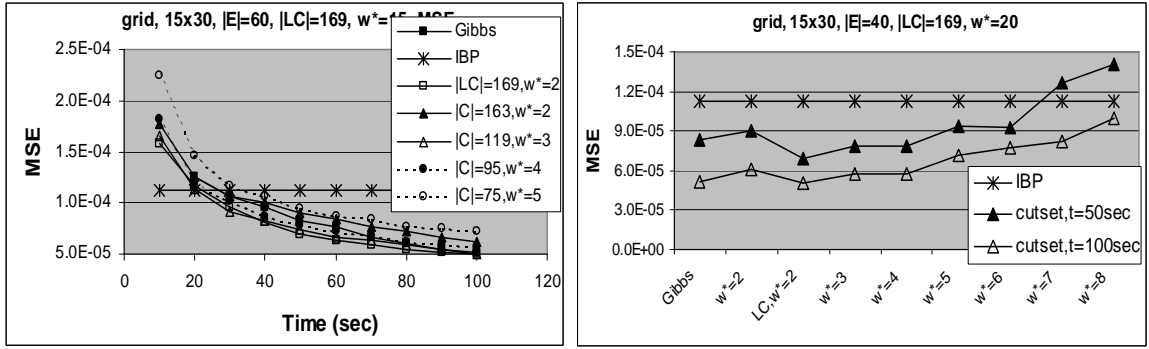

Figure 21: Random networks, 450 nodes, 10 instances. MSE as a function of the number of samples (left) and $w$ (right).

**Grid networks.** Grid networks having 450 nodes (15x30) were the only class of benchmarks where full Gibbs sampling was able to produce estimates comparable to cutset-sampling (Figure 21). With respect to accuracy, the Gibbs sampler, loop-cutset sampling, and 3-cutset sampling were the best performers and achieved similar results. Loop-cutset sampling was the fastest and most accurate among cutset sampling schemes. Still, it generated samples about 4 times more slowly compared to Gibbs sampling (Table 2) since loop-cutset is relatively large. The accuracy of loop-cutset sampling was closely followed by 2-, 3- and 4-cutset sampling slowly degrading as $w$ increased. Grid networks are an example of benchmarks with regular graph structure (that cutset sampling cannot exploit to its advantage) and small CPTs (in a two-dimensional grid network each node has at most 2 parents) where Gibbs sampling is strong.

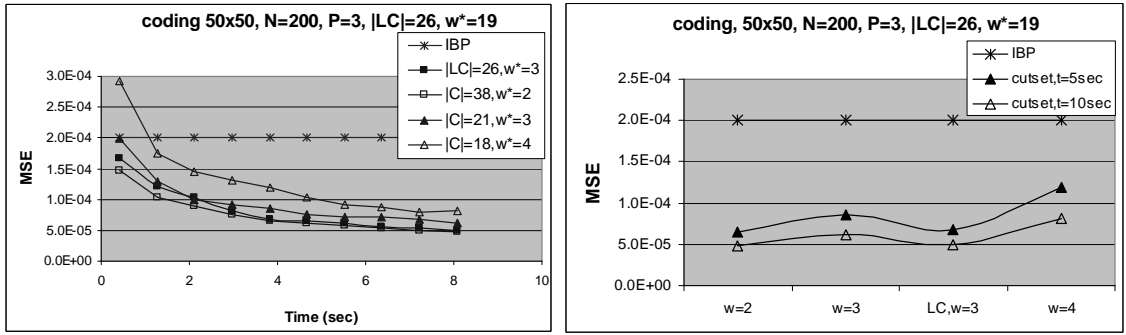

Figure 22: Coding networks, 50 code bits, 50 parity check bits, $\sigma$=0.4, 100 instances, time bound=6 minutes.

| | | Markov Chain Length $T$ | | | | | | |
|---|---|---|---|---|---|---|---|---|
| | Time | Gibbs | LC | $w$=2 | $w$=3 | $w$=4 | $w$=5 | $w$=6 |
| cpcs54 | 20 sec | 4500 | 2200 | 4000 | 2400 | 800 | 500 | - |
| cpcs179 | 40 sec | 1500 | 400 | 400 | 150 | 40 | 10 | - |
| cpcs360b | 100 sec | 2000 | 3000 | 3000 | 2000 | 800 | 500 | 200 |
| cpcs422b | 200 sec | 3000 | 20 | 2000 | 1500 | 900 | 500 | 250 |
| grid15x30 | 100 sec | 2000 | 500 | 300 | 260 | 150 | 105 | 60 |
| random | 50 sec | 2000 | 1000 | 1400 | 700 | 450 | 300 | 140 |
| 2layer | 20 sec | 200 | 700 | 900 | 320 | 150 | 75 | 40 |
| coding | 20 sec | 650 | 450 | 800 | 600 | 250 | 150 | 100 |

Table 3: Individual Markov chain length as a function of $w$. The length of each chain $M$ was adjusted for each sampling scheme for each benchmark so that the total processing time across all sampling algorithms was the same.

**Coding Networks.** The cutset sampling results for coding networks are shown in Figure 22. Here, the induced width varied from 18 to 22 allowing for exact inference. However, we additionally tested and observed that the complexity of the network grows exponentially with the number of coding bits (even after a small increase in the number of coding bits to 60 yielding a total of 240 nodes after corresponding adjustments to the number of parity-checking bits and transmitted code size, the induced width exceeds 24) while the time for each sample generation scales up linearly. We collected results for 10 networks (10 different parity check matrices) with 10 different evidence instantiations (total of 100 instances). In decoding, the Bit Error Rate (BER) is a standard error measure. However, we computed MSE over all unobserved nodes to evaluate the quality of approximate results more precisely. As expected, Gibbs sampling did not converge (because the Markov chain was non-ergodic) and was left off the charts. The charts in Figure 22 show that loop-cutset is an optimal choice for coding networks whose performance is closely followed by 2-cutset sampling. As we saw earlier, cutset sampling outperforms IBP.

## 4.5 Computing an Error Bound

Second to the issue of convergence of sampling scheme is always the problem of predicting the quality of the estimates and deciding when to stop sampling. In this section, we compare empirically the error intervals for Gibbs and cutset sampling estimates.

Gibbs sampling and cutset sampling are guaranteed to converge to the correct posterior distribution in ergodic networks. However, it is hard to estimate how many samples are needed to achieve a certain degree of convergence. It is possible to derive bounds on the absolute error based on sample variance for any sampling method if the samples are independent. In Gibbs and other MCMC methods, samples are dependent and we cannot apply the confidence interval estimate directly. In case of Gibbs sampling, we can apply the *batch means* method that is a special case of *standardized time series* method and is used by the BUGS software package (Billingsley, 1968; Geyer, 1992; Steiger & Wilson, 2001).

The main idea is to "split" a Markov chain of length $M \cdot T$ into $M$ chains of length $T$. Let $\hat{P}_m(x_i|e)$ be an estimate derived from a single chain $m \in [1, ..., M]$ of length $T$ (meaning, containing $T$ samples) as defined in equations (28)-(29). The estimates $\hat{P}_m(x|e)$ are assumed *approximately* independent for large enough $M$. Assuming that convergence conditions are satisfied and the central limit theorem holds, the $\hat{P}_m(x|e)$ is distributed according to $N(E[P(x_i|e)], \sigma^2)$ so that the posterior marginal $\hat{P}(X_i|e)$ is obtained as an average of the $M$ results obtained from each chain, namely:

$$\hat{P}(x|e) = \frac{1}{M} \sum_{m=1}^{M} \hat{P}_m(x|e) \tag{33}$$

and the sampling variance is computed as usually:

$$\sigma^2 = \frac{1}{M-1} \sum_{m=1}^{M} (\hat{P}_m(x|e) - \hat{P}(x|e))^2$$

An equivalent expression for the sampling variance is:

$$\sigma^2 = \frac{\sum_{m=1}^{M} \hat{P}_m^2(x|e) - M\hat{P}^2(x|e)}{M-1} \tag{34}$$

where $\sigma^2$ is easy to compute incrementally storing only the running sums of $\hat{P}_m(x|e)$ and $\hat{P}_m^2(x|e)$. Therefore, we can compute the confidence interval in the $100(1-\alpha)$ percentile used for random variables with normal distribution for small sampling set sizes. Namely:

$$P\left[ P(x|e) \in [\hat{P}(x|e) \pm t_{\frac{\alpha}{2},(M-1)} \sqrt{\frac{\sigma^2}{M}} \right] = 1 - \alpha \tag{35}$$

where $t_{\frac{\alpha}{2},(M-1)}$ is a table value from t distribution with $(M-1)$ degrees of freedom.

We used the batch means approach to estimate the confidence interval in the posterior marginals with one modification. Since we were working with relatively small sample sets (a few thousand samples) and the notion of "large enough" $M$ is not well defined, we restarted the chain after every $T$ samples to guarantee that the estimates $\hat{P}_m(x|e)$ were truly independent. The method of batch means only provides meaningful error estimates assuming that the samples are drawn from the stationary distribution. We assume that in our problems the chains mix fast enough so that the samples are drawn from the target distribution.

We applied this approach to estimate the error bound in the Gibbs sampler and the cutset sampler. We have computed a 90% confidence interval for the estimated posterior marginal $P(x_i|e)$ based on the sampling variance of $P_m(x_i|e)$ over 20 Markov chains as described above. We computed sampling variance $\sigma^2$ from Eq.(34) and the 90% confidence interval $\Delta_{0.9}(x_i)$ from Eq.(35) and averaged over all nodes:

$$\Delta_{0.9} = \frac{1}{N \sum_i |\mathcal{D}(X_i)|} \sum_i \sum_{x_i \in \mathcal{D}(X_i)} \Delta_{0.9}(x_i)$$

The estimated confidence interval can be too large to be practical. Thus, we compared $\Delta_{0.9}$ with the empirical average absolute error $\Delta$:

| | | **Average Error and Confidence Interval** | | | | | | |
|---|---|---|---|---|---|---|---|---|
| | | Gibbs | LC | $w$=2 | $w$=3 | $w$=4 | $w$=5 | $w$=6 |
| cpcs54 | $\Delta$ | 0.00056 | 0.00036 | 0.00030 | 0.00030 | 0.00040 | 0.00036 | 0.00067 |
| | $\Delta_{0.9}$ | 0.00119 | 0.00076 | 0.00064 | 0.00063 | 0.00098 | 0.00112 | 0.00116 |
| cpcs179 | $\Delta$ | 0.01577 | 0.00086 | 0.00074 | 0.00066 | 0.00113 | 0.00178 | - |
| | $\Delta_{0.9}$ | 0.02138 | 0.00148 | 0.00111 | 0.00164 | 0.00235 | 0.00392 | - |
| cpcs360b | $\Delta$ | 0.00051 | 0.00011 | 0.00010 | 0.00008 | 0.00014 | 0.00012 | 0.00022 |
| | $\Delta_{0.9}$ | 0.00113 | 0.00022 | 0.00023 | 0.00021 | 0.00030 | 0.00028 | 0.00046 |
| cpcs422b | $\Delta$ | 0.00055 | - | 0.00018 | 0.00020 | 0.00018 | 0.00027 | 0.00037 |
| | $\Delta_{0.9}$ | 0.00119 | - | 0.00033 | 0.00035 | 0.00043 | 0.00060 | 0.00074 |
| random | $\Delta$ | 0.00091 | 0.00039 | 0.00119 | 0.00091 | 0.00099 | 0.00109 | 0.00113 |
| | $\Delta_{0.9}$ | 0.00199 | 0.00080 | 0.00247 | 0.00205 | 0.00225 | 0.00222 | 0.00239 |
| 2layer | $\Delta$ | 0.00436 | 0.00066 | 0.00063 | 0.00082 | 0.00117 | 0.00134 | 0.00197 |
| | $\Delta_{0.9}$ | 0.00944 | 0.00145 | 0.00144 | 0.00185 | 0.00235 | 0.00302 | 0.00341 |
| coding | $\Delta$ | - | 0.00014 | 0.00019 | 0.00019 | 0.000174 | - | - |
| | $\Delta_{0.9}$ | - | 0.00030 | 0.00035 | 0.00034 | 0.000356 | - | - |
| grid15x30 | $\Delta$ | 0.00108 | 0.00099 | 0.00119 | 0.00091 | 0.00099 | 0.00109 | 0.00113 |
| | $\Delta_{0.9}$ | 0.00248 | 0.00214 | 0.00247 | 0.00205 | 0.00225 | 0.00222 | 0.00239 |

Table 4: Average absolute error $\Delta$ (measured) and estimated confidence interval $\Delta_{0.9}$ as a function of $w$ over 20 Markov Chains.

$$\Delta = \frac{1}{N \sum_i |\mathcal{D}(X_i)|} \sum_i \sum_{x_i \in \mathcal{D}(X_i)} |\hat{P}(x_i|e) - P(x_i|e)|$$

The objective of this study was to observe whether the computed confidence interval $\Delta_{0.9}$ (estimated absolute error) accurately reflects the true absolute error $\Delta$, namely, to verify that $\Delta < \Delta_{0.9}$, and if so, then investigate empirically whether confidence interval for cutset-sampling estimates will be smaller compared to Gibbs sampling as we would expect due to variance reduction.

Table 4 presents the average confidence interval and average absolute error for our benchmarks. For each benchmark, the first row of results (row $\Delta$) reports the average absolute error and the second row of results (row $\Delta_{0.9}$) reports the 90% confidence interval. Each column in Table 4 corresponds to a sampling scheme. The first column reports results for Gibbs sampling. The second column reports results for loop-cutset sampling. The remaining columns report results for $w$-cutset sampling for $w$ in range $2-6$. The loop-cutset sampling results for cpcs422b are not included due to statistically insignificant number of samples generated by loop-cutset sampling. The Gibbs sampling results for coding networks are left out because the network is not ergodic (as mentioned earlier) and Gibbs sampling does not converge.

We can see that for all the networks $\Delta < \Delta_{0.9}$ which validates our method for measuring confidence interval. In most cases the estimated confidence interval $\Delta_{0.9}$ is no more than 2-3 times the size of average error $\Delta$ and is relatively small. In case of cutset sampling, the largest confidence interval max $\Delta_{0.9} = 0.00247$ is reported in grid networks for loop-cutset sampling. Thus, the confidence interval estimate could be used as a criteria reflecting the quality of the posterior marginal estimate by the sampling algorithm in practice. Subsequently, comparing the results for Gibbs sampling and cutset sampling, we observe not only a significant reduction in the average absolute error, but also a similar reduction in the estimated confidence interval. Across all benchmarks, the estimated confidence interval of the Gibbs sampler remains $\Delta_{0.9} >$ 1E-3. At the same time, for cutset sampling we obtain $\Delta_{0.9} <$ 1E-3 in 5 out of 8 classes of networks (excluded are the cpcs179, grid, and 2-layer networks).

## 4.6 Discussion

Our empirical evaluation of the performance of cutset sampling demonstrates that, except for grid networks, sampling on a cutset usually outperforms Gibbs sampling. We show that convergence of cutset sampling in terms of number of samples dramatically improves as predicted theoretically.

The experiments clearly show that there exists a range of $w$-values where $w$-cutset sampling outperforms Gibbs sampler. The performance of $w$-cutset sampling deteriorates when increase in $w$ yields only a small reduction in the cutset size. An example is cpcs360b network where starting with $w$=4, increasing $w$ by 1 results in the reducing the sampling set by only 1 node (shown in Table 1).

We observe that the loop-cutset is a good choice of cutset sampling as long as the induced width of network $w_{LC}$ conditioned on loop-cutset is reasonably small. When $w_{LC}$ is large (as in cpcs422b), loop-cutset sampling is computationally less efficient then $w$-cutset sampling for $w < w_{LC}$.

We also showed in Section 4.3 that both Gibbs sampling and loop-cutset sampling outperform the state-of-the-art AIS-BN adaptive importance sampling method when the probability of evidence is small. Consequently, all the $w$-cutset sampling schemes in Section 4.4 that outperformed Gibbs sampler in cpcs360b and cpcs422b would also outperfrom AIS-BN.

## 5. Related Work

We mention here some related work. The idea of marginalising out some variables to improve efficiency of Gibbs sampling was first proposed by Liu et al. (1994). It was successfully applied in several special classes of Bayesian models. Kong et al. (1994) applied collapsing to the bivariate Gaussian problem with missing data. Liu (1994) defined a collapsed Gibbs sampling algorithm for finding repetitive motifs in biological sequences applies by integrating out two parameters from the model. Similarly, Gibbs sampling set is collapsed in Escobar (1994), MacEachern (1994), and Liu (1996) for learning the nonparametric Bayes problem. In all of the instances above, special relationships between problem variables have been exploited to integrate several variables out resulting in a collapsed Gibbs sampling approach. Compared to this previous research work, our contribution is in defining a generic scheme for collapsing Gibbs sampling in Bayesian networks which takes advantage of the network's graph properties and does not depend on the specific form of the relationships between variables.

Jensen et al. (1995) combined sampling and exact inference in a blocking Gibbs sampling scheme. Groups of variables were sampled simultaneously using exact inference to compute the needed conditional distributions. Their empirical results demonstrate a significant improvement in the convergence of the Gibbs sampler over time. Yet, in proposed blocking Gibbs sampling, the sample contains all variables in the network. In contrast, cutset sampling reduces the set of variables that are sampled. As noted previously, collapsing produces lower variance estimates than blocking and, therefore, cutset sampling should require fewer samples to converge.

A different combination of sampling and exact inference for join-trees was described by Koller et al. (1998) and Kjaerulff (1995). oller et al. and Kjaerulff proposed to sample the probability distribution in each cluster for computing the outgoing messages. Kjaerulff used Gibbs sampling only for large clusters to estimate the joint probability distribution $P(V_i), V_i \subset X$ in cluster $i$. The estimated $\hat{P}(V_i)$ is recorded instead of the true joint distribution to conserve memory. The motivation is that only high-probability tuples will be recorded while the remaining low-probability tuples are assumed to have probability 0. In small clusters, the exact joint distribution $P(V_i)$ is computed and recorded. However, the paper does not analyze the introduced errors or compare the performance of this scheme with standard Gibbs sampler or the exact algorithm. No analysis of error is given nor comparison with other approaches.

Koller et al. (1998) used sampling used to compute messages sent from cluster $i$ to cluster $j$ and the posterior joint distributions in a cluster-tree that contains both discrete and continuous variables. This approach subsumes the cluster-based sampling proposed by Kjaerulff (1995) and includes rigorous analysis of the error in the estimated posterior distributions. The method has difficulties with propagation of evidence. The empirical evaluation is limited to two hybrid network instances and compares the quality of the estimates to those of likelihood weighting, an instance of importance sampling that does not perform well in presence of low-probability evidence.

The effectiveness of collapsing of sampling set has been demonstrated previously in the context of Particle Filtering method for Dynamic Bayesian networks (Doucet, Andrieu, & Godsill, 2000a; Doucet, deFreitas, & Gordon, 2001; Doucet, de Freitas, Murphy, & Russell, 2000b). It was shown that sampling from a subspace combined with exact inference (Rao-Blackwellised Particle Filtering) yields a better approximation than Particle Filtering on the full set of variables. However, the objective of the study has been limited to observation of the effect in special cases where some of the variables can be integrated out easily. Our cutset sampling scheme offers a generic approach to collapsing a Gibbs sampler in any Bayesian network.

## 6. Conclusion

The paper presents the $w$-cutset sampling scheme, a general scheme for collapsing Gibbs sampler in Bayesian networks. We showed theoretically and empirically that cutset sampling improves the convergence rate and allows sampling from non-ergodic network that has ergodic subspace. By collapsing the sampling set, we reduce the dependence between samples by marginalising out some of the highly correlated variables and smoothing the sampling distributions of the remaining variables. The estimators obtained by sampling from a lower-dimensional space also have a lower sampling variance. Using the induced width $w$ as a controlling parameter, $w$-cutset sampling provides a mechanism for balancing sampling and exact inference.

We studied the power of cutset sampling when the sampling set is a loop-cutset and, more generally, when the sampling set is a $w$-cutset of the network (defined as a subset of variables such that, when instantiated, the induced width of the network is $\leq w$). Based on Rao-Blackwell theorem, cutset sampling requires fewer samples than regular sampling for convergence. Our experiments showed that this reduction in number of samples was time-wise cost-effective. We confirmed this over a range of randomly generated and real benchmarks. We also demonstrated that cutset sampling is superior to the state of the art AIS-BN importance sampling algorithm when the probability of evidence is small.

Since the size of the cutset and the correlations between the variables are two main factors contributing to the speed of convergence, $w$-cutset sampling may be optimized further with the advancement of methods for finding minimal $w$-cutset. Another promising direction for future research is to incorporate the heuristics for avoiding selecting strongly-correlated variables into a cutset since those correlations are driving factors in the speed of convergence of Gibbs sampling. Alternatively, we could combine sample collapsing with blocking.

In summary, $w$-cutset sampling scheme is a simple yet powerful extension of sampling in Bayesian networks that is likely to dominate regular sampling for any sampling method. While we focused on Gibbs sampling with better convergence characteristics, other sampling schemes can be implemented with the cutset sampling principle. In particular, it was adapted for use with likelihood weighting (Bidyuk & Dechter, 2006).